\documentclass{article}
\PassOptionsToPackage{numbers, compress}{natbib}
\usepackage[main, final]{neurips_2026}

\usepackage[utf8]{inputenc}
\usepackage[T1]{fontenc}

\usepackage{hyperref}
\usepackage{url}

\usepackage{booktabs}
\usepackage{multirow}

\usepackage{amsmath}
\usepackage{amsfonts}
\usepackage{amssymb}
\usepackage{mathtools}

\usepackage{graphicx}
\usepackage{subcaption}
\usepackage{wrapfig}

\usepackage{nicefrac}
\usepackage{microtype}

\usepackage{xcolor}

\usepackage{algorithm}
\usepackage{algorithmic}

\usepackage{enumitem}
\usepackage[capitalise,noabbrev]{cleveref}

\usepackage{tabularx}
\usepackage[table]{xcolor}

\newcommand{\R}{\mathbb{R}}

\newcommand{\E}{\mathbb{E}}
\newcommand{\bx}{\mathbf{x}}

\newcommand{\br}{\mathbf{r}}

\newcommand{\bp}{\mathbf{p}}

\newcommand{\bh}{\mathbf{h}}

\graphicspath{{figs/}}

\title{PixelDense: Dense Prediction as Representation Alignment for Pixel Diffusion}

\author{%
  Lehan Yang$^{1}$ \quad
  Daiqing Qi$^{1}$ \quad
  Wenhao Zhang$^{1}$ \quad
  Avery Li$^{1}$ \quad
  Yiqing Yang$^{1}$ \quad
  Yifan Li$^{2}$ \\
  \bfseries %
  Yu Kong$^{2}$ \quad
  Haitian Zheng$^{3}$ \quad
  Zhifei Zhang$^{3}$ \quad
  Zhe Lin$^{3}$ \quad
  Varun Jampani$^{4}$ \quad
  Sheng Li$^{1}$ \\[4pt]
  $^{1}$University of Virginia \qquad
  $^{2}$Michigan State University \qquad
  $^{3}$Adobe \qquad
  $^{4}$Arcade AI \\
  \texttt{lehan.yang@virginia.edu}
}
\begin{document}

\maketitle

\vspace{-4mm}
\begin{center}
  \includegraphics[width=\textwidth]{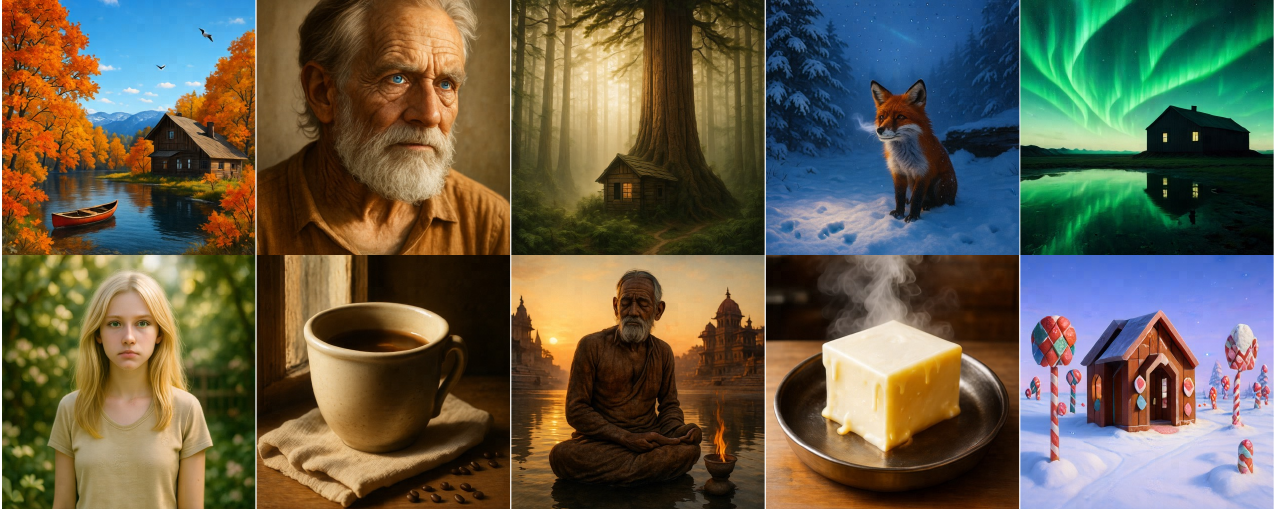}
  \captionof{figure}{\textbf{Qualitative text-to-image results of PixelDense.} PixelDense generates diverse $512 \times 512$ samples with strong compositional fidelity, including object count, attribute binding, spatial relations, and geometry-aware scene layout.}
  \label{fig:teaser}
\end{center}

\begin{abstract}
  Representation alignment (REPA) accelerates diffusion transformer training, but its alignment targets are almost exclusively semantic encoders such as DINOv2 and CLIP.
Recent analysis points to spatial structure, not global semantics, as the carrier of the alignment effect, yet dense-prediction foundation models trained to predict that structure remain overlooked as REPA targets.
In pixel-space diffusion, SAM2, Depth Anything v2, and Metric3D v2 each outperform the DINOv2-only GenEval baseline, with the two geometric teachers leading the segmentation teacher.
A flat sum of all four teachers, however, lands below the best single geometric teacher, as semantic and geometric gradients compete for one denoiser projection.
We introduce \emph{PixelDense}, which routes DINOv2 and SAM2 through a semantic projection stream, routes Depth Anything v2 and Metric3D v2 through a geometric projection stream, and adds a weight-space orthogonality penalty that keeps the two streams in disjoint subspaces.
All four teachers are frozen during training and dropped at inference.
Applied to PixelGen and DeCo with a single recipe, PixelDense improves GenEval, DPG-Bench, and HPS v2.1, raises PixelGen-XXL's GenEval Overall from $0.7927$ to $0.8093$, and beats every single-teacher and unfactored multi-teacher variant.
In partial-noise reconstruction, independent panoptic, depth, and surface-normal probes show up to $53.1\%$ PQ gain and $36.0\%$ depth AbsRel reduction at $\tau=0.5$ across COCO and Flickr30K.
From random initialization, PixelDense also reaches the baseline's peak GenEval $1.23\times$ faster.
In SDEdit editing on PIE-Bench, PixelDense keeps more of the source background and layout at every edit strength, raising background PSNR by up to $2.2$ dB.
\end{abstract}

\section{Introduction}
\label{sec:intro}
\vspace{-2mm}

Dense-prediction foundation models can now extract increasingly rich spatial structure from a single RGB image: object-level semantic features from DINOv2~\citep{oquab2023dinov2}, instance and region boundaries from SAM2~\citep{kirillov2023segment,ravi2024sam}, monocular depth from Depth Anything v2~\citep{yang2024depth,yang2024depthv2}, and metric surface normals from Metric3D v2~\citep{hu2024metric3d}. 
Together, they capture object semantics, boundaries, depth, and surface orientation, jointly providing semantic and geometric constraints on what a realistic image should look like.
We therefore ask whether such semantic and geometric knowledge can directly supervise image generation during training, without requiring dense-prediction outputs at inference.

The recent REPresentation Alignment (REPA)~\citep{yu2025representation} provides a direct way to use a frozen encoder to supervise diffusion training. 
By matching intermediate denoiser features to frozen encoder features, REPA accelerates DiT training by an order of magnitude without changing inference. 
However, REPA-style alignment has primarily relied on semantic encoders such as DINOv2 and CLIP~\citep{radford2021learning}. 
iREPA~\citep{singh2025matters} shows that REPA-style alignment benefits primarily from the spatial structure rather than the global semantics of the teacher encoder.
This suggests that dense-prediction foundation models may provide stronger structural guidance for diffusion models, as they are explicitly trained to capture spatial structures.
Although video diffusion has explored geometric~\citep{wu2025geometry} and multi-encoder~\citep{lee2025improving} alignment for temporal consistency, it remains unclear how to use
semantic and geometric encoders as structural constraints for realistic image synthesis.

We instantiate this idea in the setting of pixel-space diffusion, where the denoiser tokens lie on the RGB patch grid that dense teachers consume, so each token aligns directly with the teacher feature. On a PixelGen~\citep{ma2026pixelgen} text-to-image backbone fine-tuned at $512{\times}512$, we show that each of SAM2, Depth Anything v2, and Metric3D v2 improves GenEval Overall over the DINOv2-only baseline when used as the added alignment target, and the two geometric teachers outperform the segmentation teacher. This matches iREPA's finding that spatial structure carries the alignment signal.

However, dense teachers do not compose under a flat sum. 
A four-way summed REPA loss across DINOv2, SAM2, Depth Anything v2, and Metric3D v2 through a single denoiser projection lands below the best single geometric teacher, diluting the fine-grained structure that motivated adding more teachers in the first place. 
The cause is that semantic and geometric teachers reward incompatible invariances. 
Semantic encoders push toward viewpoint invariance and instance identity, while geometric encoders push toward metric layout and surface orientation. 
When both groups of gradients share a single feature basis, the denoiser cannot allocate distinct subspaces to distinct kinds of structure, and the teachers compete instead of compose. 
The core challenge is therefore representation routing. 
Semantic and geometric teachers must supervise the same denoiser through separate subspaces while the RGB sampler stays unchanged.

Dense perception offers richer alignment supervision than semantic encoders alone, but only if the alignment objective respects the semantic and geometric structure of the teacher bank. We propose \emph{PixelDense}, a factored dense-alignment framework that provides this structure. It decomposes the multi-teacher objective into two disjoint streams: a semantic stream that carries DINOv2 and SAM2 through a shared semantic projection bottleneck, and a geometric stream that carries Depth Anything v2 and Metric3D through a separate geometric projection bottleneck. A weight-space orthogonality regularizer on the two bottlenecks' input projections keeps the denoiser from collapsing them into a common subspace, so the two streams specialize rather than average. The teachers remain frozen and are dropped at inference, so PixelDense adds no cost to sampling.

PixelDense is not tied to a particular pixel-space backbone. We instantiate it on PixelGen and on DeCo~\citep{ma2025deco}, two pixel diffusion transformers with different loss recipes, and the same alignment recipe transfers without modification. On PixelGen at $512{\times}512$, full PixelDense raises GenEval Overall over the DINOv2-only baseline from $0.7927$ to $0.8093$, a $2.09\%$ relative gain, and beats every single-teacher and unfactored multi-teacher variant we compare against. It also preserves source-image structure more faithfully than PixelGen in partial-noise reconstruction. Under standalone segmentation, depth, and surface-normal probes, PixelDense has better scores in every COCO and Flickr30K setting in \Cref{tab:geometry_reconstruction}. At $\tau=0.5$, panoptic PQ rises by $35.3\%$ against COCO panoptic annotations and by $53.1\%$ against Flickr30K pseudo labels. Depth AbsRel falls by up to $36.0\%$. On DeCo, PixelDense also improves over the baseline.
The same structure preservation carries over to image editing: with SDEdit on PIE-Bench, PixelDense keeps more of the source background and layout at every edit strength.

The main contributions of this work are summarized as follows:
\vspace{-2mm}
\begin{itemize}
    \item We introduce \emph{PixelDense}, a dense-perception representation-alignment framework for pixel-space diffusion. PixelDense uses semantic, segmentation, depth, and surface-normal models as training-time teachers while retaining the original RGB generator for inference.
    \item We propose a semantic--geometric factorization for dense teacher banks. PixelDense routes DINOv2 and SAM2 through a semantic projection stream, routes Depth Anything v2 and Metric3D v2 through a geometric projection stream, and uses weight space orthogonality to keep the two forms of supervision from collapsing into the same denoiser subspace.
    \item We identify a negative-transfer effect in dense multi-teacher alignment. Although each dense teacher improves over the DINOv2-only baseline, simply summing semantic and geometric REPA losses forces incompatible invariances through a shared projection and degrades the combined signal. This reframes teacher composition as a representation-routing problem.
    \item Extensive experiments demonstrate the effectiveness and transferability of PixelDense. On PixelGen and DeCo, PixelDense improves compositional generation over matched fine-tunes, single-teacher alignment, and unfactored multi-teacher baselines. In partial-noise reconstruction, standalone segmentation, depth, and surface-normal estimators show stronger structure preservation on COCO and Flickr30K, and this carries over to SDEdit image editing on PIE-Bench.
\end{itemize}

\section{Related Work}
\label{sec:related_work}

\noindent\textbf{Diffusion models for image generation.}
Diffusion models~\citep{ho2020denoising, song2021denoising} and flow matching~\citep{lipman2023flow, liu2023flow} now form a common image synthesis framework, and classifier-free guidance~\citep{ho2022classifier} has become a standard conditioning tool. Latent diffusion~\citep{rombach2022high} made compressed VAE space the default for efficient generation, while transformer denoisers including DiT~\citep{peebles2023scalable}, U-ViT~\citep{bao2023all}, SiT~\citep{ma2024sit}, MDT~\citep{gao2023masked}, and DDT~\citep{wang2025ddt} improved the backbone. Pixel-space methods remove the VAE bottleneck and its reconstruction artifacts. Simple Diffusion~\citep{hoogeboom2023simple}, PixelFlow~\citep{chen2025pixelflow}, and self-supervised pretraining~\citep{lei2025there} show that direct raw-pixel training is viable, and JiT~\citep{li2025back}, DeCo~\citep{ma2025deco}, PixelDiT~\citep{yu2025pixeldit}, PixelGen~\citep{ma2026pixelgen}, and Latent Forcing~\citep{baade2026latent} further improve prediction targets, frequency treatment, architectures, perceptual losses, or trajectory design. These methods strengthen the denoisers, but their supervision remains an RGB reconstruction target, which leaves the semantic and geometric structure of the image implicit. We build on this pixel-space line and ask what additional training signal can supervise the denoiser.

\noindent\textbf{Representation alignment for diffusion models.}
REPA~\citep{yu2025representation} answers this question by matching intermediate denoiser features to a frozen encoder such as DINOv2~\citep{oquab2023dinov2}, accelerating DiT training with no inference cost. iREPA~\citep{singh2025matters} further identified spatial structure as the active ingredient of the alignment signal, which reframes alignment as a question about which encoder exposes that structure most directly. Follow-up work extends the recipe through VAE tuning (REPA-E~\citep{leng2025repa}), hierarchical alignment (SARA~\citep{chen2025sara}), time-limited objectives (HASTE~\citep{wang2025repa}, ERW~\citep{liu2025efficient}, curriculum decay~\citep{wang2025learning}), and entanglement or self-alignment variants (REG~\citep{wu2025representation}, LayerSync~\citep{haghighi2025layersync}), while PixelREPA~\citep{shin2026representation} and V-Co~\citep{lin2026v} study pixel-space collapse and co-denoising. Closer to us, Align4Gen~\citep{lee2025improving} and Geometry Forcing~\citep{wu2025geometry} bring multi-encoder geometry-aware alignment to video diffusion, but they target temporal consistency and sum teacher signals without addressing the conflict between semantic and geometric supervision. Image-generation alignment still defaults to single semantic teachers such as DINOv2, CLIP~\citep{radford2021learning}, and SigLIP~\citep{zhai2023sigmoid}, leaving open which encoders actually carry the spatial structure iREPA points to.

\noindent\textbf{Dense prediction meets generation.}
Dense prediction foundation models, including SAM~\citep{kirillov2023segment} and SAM2~\citep{ravi2024sam} for segmentation, Depth Anything~\citep{yang2024depth} and Depth Anything V2~\citep{yang2024depthv2} for depth, and Metric3D v2~\citep{hu2024metric3d} for depth and surface normals, are trained explicitly to expose this spatial structure from a single RGB image. Most prior work explores the reverse direction. Marigold~\citep{ke2024repurposing}, GeoWizard~\citep{fu2024geowizard}, Lotus~\citep{he2024lotus}, GenPercept~\citep{xu2024matters}, VPD~\citep{zhao2023unleashing}, and ODISE~\citep{xu2023open} repurpose diffusion features for dense perception, while UniGS~\citep{qi2024unigs}, SemFlow~\citep{wang2024semflow}, UDPDiff~\citep{yang2025unified}, and Diff-2-in-1~\citep{zheng2024diff} couple image and dense outputs in unified generators. These models still emit dense predictions or target perception gains. We take the opposite direction and use these frozen discriminative encoders as alignment teachers for a pixel-space generator, routing semantic and geometric supervision through separate streams so the two forms of structure specialize rather than average, and the dense teachers are dropped at inference.

\section{Method}
\label{sec:method}

\begin{figure}[t]
    \centering
    \includegraphics[width=0.95\textwidth]{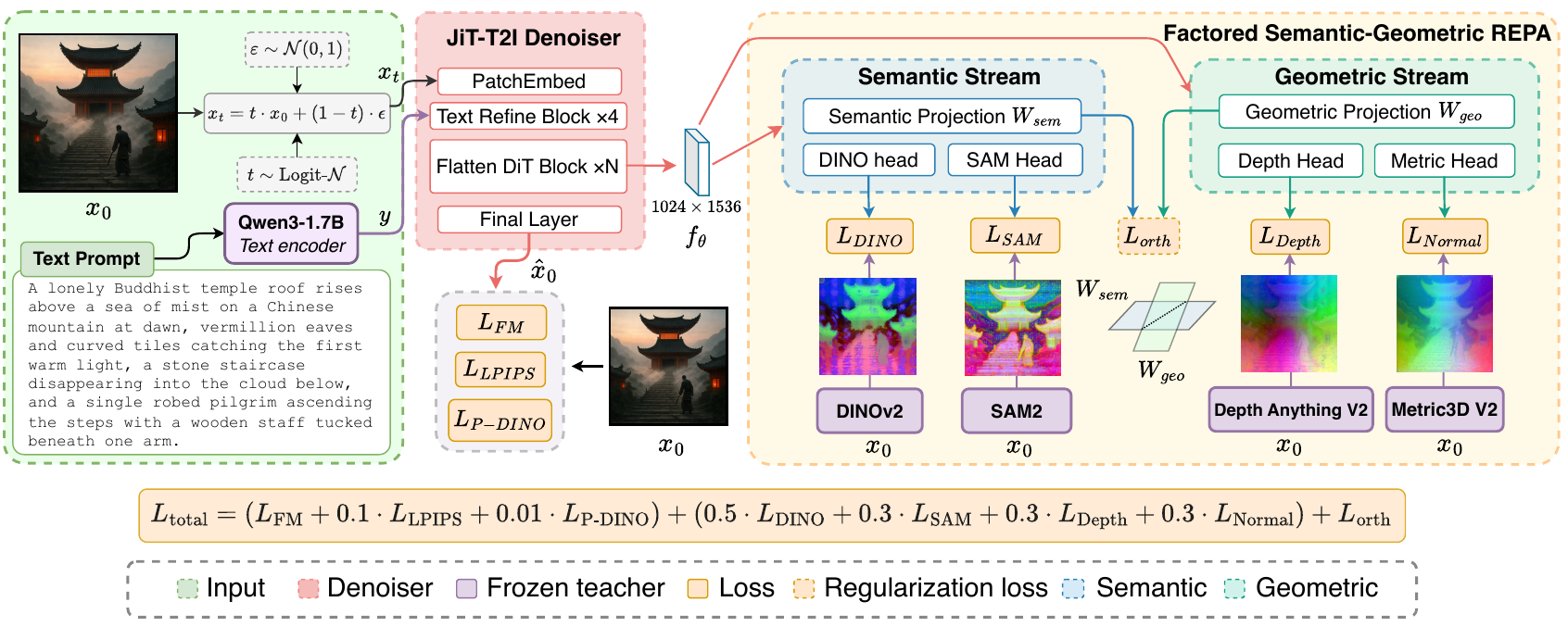}
    \vspace{-2mm}
    \caption{\textbf{PixelDense factors dense-prediction supervision into a semantic stream and a geometric stream.} The pixel diffusion denoiser $f_\theta$ takes the noisy image $\bx_t$ and a Qwen3 text embedding $y$, and reads its block-$8$ token state into two parallel projections, a $D_s = 768$ semantic projection $W_\text{sem}$ aligned to frozen DINOv2 and SAM2 features, and a $D_g = 1024$ geometric projection $W_\text{geo}$ aligned to frozen Depth Anything v2 and Metric3D v2 features. A weight-space orthogonality penalty $\mathcal{L}_\text{orth}$ on the rows of $W_\text{sem}$ and $W_\text{geo}$ keeps the two streams from collapsing into a shared subspace of the denoiser feature. Only the denoiser, the projections, and the per-teacher heads receive gradients. The four dense teachers are frozen during training and dropped at inference, so the deployed model is the original three-channel RGB generator with its original sampler.}
    \label{fig:framework}
    \vspace{-4mm}
\end{figure}

Based on a pixel-space diffusion transformer~\citep{ma2026pixelgen, ma2025deco}, the proposed PixelDense framework distills semantic and geometric perception into the denoiser through factored representation alignment. \Cref{fig:framework} gives an overview of PixelDense.
An intermediate denoiser token state is projected into two parallel streams: a semantic stream aligned to DINOv2 and SAM2 features, and a geometric stream aligned to Depth Anything v2 and Metric3D v2 features.
A weight-space orthogonality penalty on the streams' input projections keeps the two streams from collapsing into a shared subspace of the denoiser feature.
All four teachers are frozen during training and dropped at inference.

\subsection{Pixel Diffusion with Single-Teacher REPA}
\label{sec:method_setup}

We work with a pixel-space diffusion transformer $f_\theta$ that maps a noised image $\bx_t \in \R^{H \times W \times 3}$ and a text embedding $y$ to a clean pixel prediction $\hat{\bx}_0 = f_\theta(\bx_t,t,y)$. The denoiser is trained with a flow-matching velocity loss $\mathcal{L}_\text{FM}$ on $\hat{\bx}_0$. The clean image, scaled to $[0,1]$, denoted $\br$, is the input domain of the frozen teachers.

\noindent\textbf{Denoiser features.}
At an intermediate transformer block $\ell$, we read the visual-stream token state
$ \bh_\ell \in \R^{N \times d_h}$,
where $N$ is the number of patch tokens and $d_h$ is the hidden width. Each token $\bh_{\ell,n}$ is anchored to a patch on the same image grid that the dense teachers consume, so its alignment target is a single teacher-feature vector at the matching spatial location. We set $\ell=8$ in a $16$-block denoiser.

\noindent\textbf{Single-teacher REPA.}
For any teacher $k \in \mathcal{K}$, a frozen teacher encoder $T_k$ maps the clean image to a token-aligned feature grid $T_k(\br) \in \R^{N \times d_k}$. With a tokenwise projector $P_k:\R^{d_h}\to\R^{d_k}$, REPA aligns the projected denoiser feature to the teacher feature with a cosine loss,
\begin{equation}
\mathcal{L}_k(\theta, P_k)
=
\E_{\bx_0,\varepsilon,t}\!\left[
\frac{1}{N}\sum_{n=1}^{N}
\ell_\text{cos}\!\left(P_k(\bh_{\ell,n}),\, T_k(\br)_n\right)
\right],
\quad
\ell_\text{cos}(a,b) = 1 - \frac{\langle a,b\rangle}{\|a\|\,\|b\|},
\label{eq:single_repa}
\end{equation}
where the teacher receives no gradient. This is the alignment loss used by REPA~\citep{yu2025representation}, iREPA~\citep{singh2025matters} and related follow-up works, which typically align to a single semantic teacher such as DINOv2.

\subsection{Why a Single Alignment Stream Cannot Carry Dense Perception}
\label{sec:method_failure}

We use a teacher bank $\mathcal{K} = \{\text{DINO},\text{SAM},\text{DA2},\text{M3D}\}$ that combines semantic and geometric supervision. DINOv2~\citep{oquab2023dinov2} and SAM2~\citep{ravi2024sam} capture object identity, regions, and boundaries, whereas Depth Anything v2~\citep{yang2024depthv2} and Metric3D v2~\citep{hu2024metric3d} capture 3D layout, depth, and surface orientation. As illustrated in \Cref{fig:teacher_feature_viz} of \Cref{sec:appendix_teacher_viz}, the two groups provide complementary spatial cues for the same image, motivating an alignment objective that preserves this semantic--geometric split.

\noindent\textbf{Unfactored multi-teacher REPA.}
The natural extension of \Cref{eq:single_repa} sums the per-teacher losses,
$ \mathcal{L}_\text{unfactored} = \sum_{k\in\mathcal{K}} \lambda_k\,\mathcal{L}_k(\theta, P_k),$
with independent projector $P_k$ for each teacher. Although these heads decouple the teachers' target spaces, all losses still update through the same denoiser features $\bh_{\ell,n}$. Thus, the denoiser must satisfy four supervision signals through a single input feature space.

\noindent\textbf{Gradient interference.}
The four signals need not point in compatible directions. 
Semantic encoders reward viewpoint invariance and instance identity, whereas geometric encoders reward metric layout and surface orientation. 
With a shared denoiser feature, the gradients pulling $\bh_{\ell,n}$ toward semantic identity interfere with those pulling it toward depth and normal layout. 
Dense perception offers richer alignment supervision than a single semantic teacher, but only if the alignment objective respects the semantic and geometric structure of the teacher bank.

\subsection{PixelDense: Factored Streams with Orthogonal Subspaces}
\label{sec:method_pixeldense}

As show in \Cref{fig:framework}, PixelDense replaces the single denoiser-side projector with two parallel projection streams, one for each REPA teacher group, and constrains their input maps to occupy orthogonal subspaces of $\R^{d_h}$.

\noindent\textbf{Two streams.}
A semantic stream $\pi_s$ and a geometric stream $\pi_g$ read the same denoiser token,
\begin{equation}
\pi_s(\bh_{\ell,n}) = W_\text{sem}^{(2)}\,\sigma\!\left(\mathrm{LN}(W_\text{sem}\,\bh_{\ell,n})\right),
\qquad
\pi_g(\bh_{\ell,n}) = W_\text{geo}^{(2)}\,\sigma\!\left(\mathrm{LN}(W_\text{geo}\,\bh_{\ell,n})\right),
\label{eq:streams}
\end{equation}
with $W_\text{sem}\in\R^{D_s\times d_h}$, $W_\text{geo}\in\R^{D_g\times d_h}$, SiLU nonlinearity $\sigma$, and channel-wise LayerNorm. We set $D_s=768$ and $D_g=1024$ to match the native widths of the semantic and geometric teacher groups. Four light teacher heads then map each stream output to its teachers' feature widths,
\begin{equation}
\begin{aligned}
\bp_{\text{DINO},n} &= g_\text{DINO}(\pi_s(\bh_{\ell,n})), &
\bp_{\text{SAM},n}  &= g_\text{SAM}(\pi_s(\bh_{\ell,n})), \\
\bp_{\text{DA2},n}  &= g_\text{DA2}(\pi_g(\bh_{\ell,n})), &
\bp_{\text{M3D},n}  &= g_\text{M3D}(\pi_g(\bh_{\ell,n})).
\end{aligned}
\label{eq:teacher_heads}
\end{equation}
Semantic teachers therefore share the semantic bottleneck $\pi_s$, geometric teachers share the geometric bottleneck $\pi_g$,
avoiding a single shared input projection for all four signals.

\noindent\textbf{Orthogonality between streams.}
Two streams alone do not guarantee specialization, because $\pi_s$ and $\pi_g$ both read $\bh_\ell$. If $W_\text{sem}$ and $W_\text{geo}$ learn the same row subspace of $\R^{d_h}$, the streams compete for the same denoiser coordinates. PixelDense adds a weight-space penalty on the input-facing linear maps,
\begin{equation}
\mathcal{L}_\text{orth}(W_\text{sem}, W_\text{geo})
=
\frac{1}{D_s D_g}\,
\bigl\|\bar{W}_\text{sem}\,\bar{W}_\text{geo}^{\top}\bigr\|_1,
\qquad
\bar{W}_{i,:} = \frac{W_{i,:}}{\|W_{i,:}\|_2},
\label{eq:orth}
\end{equation}
the mean absolute cosine between every semantic row and geometric row, where each row of $W$ is unit-length normalized to form $\bar W$. The penalty uses only projector weights. Therefore it is data-free and it adds no inference costs. When $\mathcal{L}_\text{orth} = 0$, every row of $W_\text{sem}$ is orthogonal to every row of $W_\text{geo}$, so semantic and geometric gradients enter $\bh_\ell$ through orthogonal subspaces at the first projection layer. The full nonlinear streams are not globally constrained to be orthogonal, but the penalty discourages the two streams from reading the same input directions at their most direct projection.

\noindent\textbf{Alignment loss.}
For each teacher, the factored cosine loss is
\begin{equation}
\mathcal{L}_k^\text{F}
=
\E_{\bx_0,\varepsilon,t}\!\left[
\frac{1}{N}\sum_{n=1}^{N}
\ell_\text{cos}\!\left(\bp_{k,n},\, T_k(\br)_n\right)
\right],
\label{eq:factored_teacher_loss}
\end{equation}
which replaces $P_k(\bh_{\ell,n})$ in \Cref{eq:single_repa} with the corresponding factored projection $\bp_{k,n}$. %
\begin{equation}
\mathcal{L}_\text{align}
=
\bigl(
0.5\,\mathcal{L}_\text{DINO}^\text{F}
+0.3\,\mathcal{L}_\text{SAM}^\text{F}
+0.3\,\mathcal{L}_\text{DA2}^\text{F}
+0.3\,\mathcal{L}_\text{M3D}^\text{F}
\bigr),
\quad
\label{eq:align}
\end{equation}

\noindent\textbf{Full objective.}
PixelDense trains the denoiser parameters $\theta$ and the alignment-side parameters $\Phi=\{\pi_s,\pi_g,g_\text{DINO},g_\text{SAM},g_\text{DA2},g_\text{M3D}\}$ under
\begin{equation}
\mathcal{L}(\theta,\Phi)
=
\mathcal{L}_\text{FM}(\theta)
+\mathcal{L}_\text{align}(\theta,\Phi)
+\lambda_\text{orth}\,\mathcal{L}_\text{orth}(W_\text{sem},W_\text{geo})
+\mathcal{L}_\text{aux}(\theta),
\label{eq:total}
\end{equation}
with $\lambda_\text{orth}=0.01$ and $\mathcal{L}_\text{aux}$ inherited from LPIPS and Perceptual-DINO loss~\cite{ma2026pixelgen}. The teachers and the parameters in $\Phi$ are dropped after training. The deployed model is the original three-channel RGB generator with its original sampler.

\section{Experiments}
\label{sec:experiments}

We first introduce the experimental setting: pixel-space backbones, dense teachers, fine-tuning recipe, datasets, metrics, and baselines. We then evaluate PixelDense on PixelGen-XXL~\citep{ma2026pixelgen} and test whether partial-noise reconstructions preserve scene geometry under independent panoptic, depth, and surface-normal predictors. Next, we analyze how each dense teacher contributes to generation quality, study multi-teacher composition, and ablate the semantic--geometric factorization, projection width, alignment block, alignment-loss weights, and orthogonality regularizer. Finally, we examine training from scratch, transfer the same recipe to DeCo~\citep{ma2025deco} without retuning, test the trained model on image editing, and provide qualitative visualization under matched prompts and samplers.

\subsection{Setup}
\label{sec:exp_setup}

Our main backbone is the 1.1B PixelGen-XXL~\citep{ma2026pixelgen}. We also finetune DeCo~\citep{ma2025deco} for testing the generality. 
Both models are pretrained with DINOv2 REPA~\citep{yu2025representation}. 
During fine-tuning, this DINOv2 signal remains active. Because both backbones already retain the original DINOv2 REPA signal during fine-tuning, our primary comparison baseline is therefore a single teacher DINOv2 baseline. 
We conduct all experiments on $2 \times$ NVIDIA H200 GPUs. 
Each GPU uses a batch size of $16$ with gradient accumulation of $8$, giving an effective total batch size of $256$. 
We train for $10{,}000$ optimizer steps with AdamW and a learning rate of $1\text{e-}5$. 
At inference, we use classifier-free guidance of $4.0$.

We use four frozen dense prediction perception models as alignment targets.
DINOv2-B/14~\citep{oquab2023dinov2} provides object-level semantic features.
SAM2 Hiera-L~\citep{ravi2024sam} provides segmentation aware boundary features from the last image-encoder layer. 
Depth Anything v2 ViT-L~\citep{yang2024depthv2} provides monocular depth features from layer $17$. 
Metric3D v2 ViT-L~\citep{hu2024metric3d} provides surface normal features from layer $17$. 
We align all teacher signals at block $8$ of the diffusion denoiser. 
All four teachers remain frozen during training and are dropped at inference.

The finetuning dataset follows the third stage of PixelGen training and uses BLIP3-o-60K~\citep{chen2025blip3}. 
We report results on GenEval~\citep{ghosh2023geneval}, DPG-Bench~\citep{hu2024ella}, and HPS v2.1~\citep{wu2023human}. 
We compare with pixel-space methods, including PixelDiT~\citep{yu2025pixeldit}, PixelFlow~\citep{chen2025pixelflow}, PixelGen~\citep{ma2026pixelgen}, and DeCo~\citep{ma2025deco}. 
We also include latent-space reference methods, SDXL~\citep{podell2023sdxl} and SD3~\citep{esser2024scaling}.

\subsection{Results}
\label{sec:exp_main}

As shown in \Cref{tab:main_compare}, PixelDense improves over the matched finetune of PixelGen across GenEval Overall, DPG-Bench, and HPS v2.1. 
It raises GenEval Overall from $0.7927$ to $0.8093$, a $2.09\%$ relative gain over the baseline. 
The gains land on the compositional axes that touch object boundaries, depth layout, and surface geometry, which are the axes that dense-prediction supervision is designed to lift. 
Compared to the listed pixel-space backbones at this scale, PixelDense reaches the strongest GenEval Overall. 
The dense teachers are removed at test time, so inference speed is not changed.

\begin{table}[t]
\centering
\caption{\textbf{PixelDense lifts compositional alignment over the matched PixelGen fine-tune at no inference-time cost.} GenEval~\citep{ghosh2023geneval}, DPG-Bench~\citep{hu2024ella}, and HPS v2.1~\citep{wu2023human} on the released PixelGen-XXL~\citep{ma2026pixelgen} backbone at $512 \times 512$. Our rows use the $10{,}000$-step, effective-batch-$256$, $2\times$H200 recipe of \Cref{sec:exp_setup}. Higher is better for every metric. Bold marks PixelDense entries that improve over the matched PixelGen fine-tune.}
\label{tab:main_compare}
\resizebox{\textwidth}{!}{%
\begin{tabular}{l l c | c c c c c c c | c | c}
\toprule
\multirow{2}{*}{Model} & \multirow{2}{*}{Type} & \multirow{2}{*}{\#Params} & \multicolumn{7}{c|}{GenEval $\uparrow$} & \multirow{2}{*}{DPG $\uparrow$} & \multirow{2}{*}{HPS $\uparrow$} \\
 & & & Sing. Obj. & Two Obj. & Count. & Colors & Attr. & Pos. & Overall & & \\
\midrule
SDXL~\citep{podell2023sdxl}                 & latent & 2.6B & 98.00 & 74.00 & 39.00 & 85.00 & 23.00 & 15.00 & 0.55 & 74.7 & -- \\
SD3~\citep{esser2024scaling}                & latent & 8B   & 98.00 & 84.00 & 66.00 & 74.00 & 43.00 & 40.00 & 0.68 & -- & -- \\
\midrule
PixelDiT (1024 res)~\citep{yu2025pixeldit}             & pixel & 1.3B & 100.00 & 94.00 & 70.00 & 90.00 & 65.00 & 53.00 & 0.78 & 83.7 & 0.250 \\
PixelFlow~\citep{chen2025pixelflow}         & pixel & 0.9B & -- & -- & -- & -- & -- & -- & 0.60 & 77.9 & -- \\
PixelGen~\citep{ma2026pixelgen}             & pixel & 1.1B & 99.00 & 88.00 & 59.00 & 90.00 & 70.00 & 70.00 & 0.79 & - & 0.281 \\
\midrule
PixelGen + fine-tune                        & pixel & 1.1B & \textbf{99.38} & 88.89 & 57.50 & 89.36 & 69.00 & 71.50 & 0.7927 & 78.7 & 0.280 \\
\textbf{PixelDense}                         & pixel & 1.1B & \textbf{99.38} & \textbf{89.39} & \textbf{58.75} & \textbf{93.09} & \textbf{70.50} & \textbf{74.50} & \textbf{0.8093} & \textbf{78.9} & \textbf{0.282} \\
\bottomrule
\end{tabular}%
}
\end{table}

\begin{figure}[t]
    \centering
    \includegraphics[width=\textwidth]{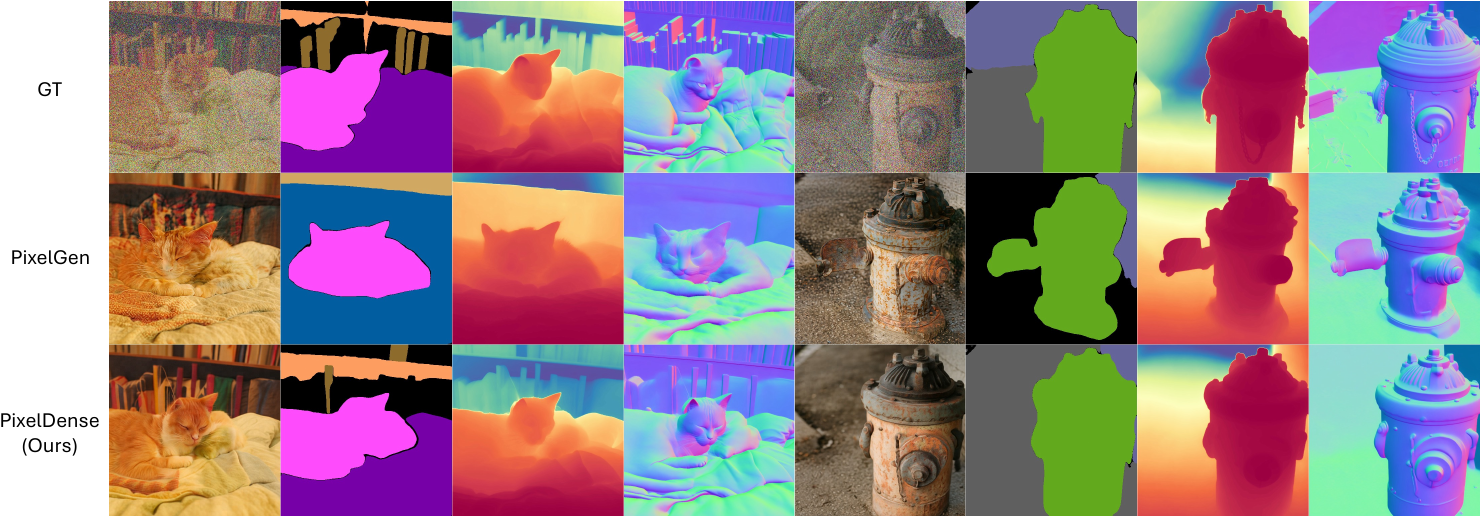}
    \caption{\textbf{Qualitative geometry preservation comparison between PixelGen and PixelDense.} We show noised/reconstructed images and pseudo labels for segmentation, depth, and normals. PixelGen shows more ghosting and structural distortions, while PixelDense better preserves GT geometry.}
    \label{fig:geometry_preservation}
\end{figure}

\noindent\textbf{Geometry Preservation under Partial-Noise Reconstruction.}
Text-to-image metrics score prompt alignment, compositional correctness, and human aesthetic preference, but they are agnostic to whether a generation preserves the geometry of a specific input scene. We therefore add a partial-noise reconstruction probe: given an input image, we corrupt it at noise levels $\tau \in \{0.5, 0.7, 0.9\}$ and feed the noised image with its caption to the model. To avoid evaluating with the same models used during training, we score reconstructions with three off-the-shelf probes that are independent of the REPA teachers. For panoptic agreement (PQ~\citep{kirillov2019panoptic} and mIoU), we use COCO-trained OneFormer~\citep{jain2023oneformer} with two backbones, Swin-L~\citep{liu2021swin} and DiNAT-L~\citep{hassani2022dilated}. For depth and surface normals, we use Marigold~\citep{ke2024repurposing}. We evaluate on COCO 2017~\citep{lin2014microsoft} val against both ground-truth panoptic and predictor-on-original pseudo references, and on Flickr30K~\citep{plummer2015flickr30k} against pseudo references only.

As shown in \Cref{tab:geometry_reconstruction}, PixelDense gives stronger structural agreement than the matched PixelGen baseline across every dataset, reference type, probe, and noise level. The gap is largest at $\tau=0.5$ and narrows at $\tau=0.9$, where severe noise removes most of the input structure for both models. Flickr30K panoptic scores are lower than COCO because OneFormer uses the COCO vocabulary; depth and normal trends remain consistent across datasets. \Cref{fig:geometry_preservation} shows the same trend qualitatively: PixelDense reconstructions preserve object boundaries, depth ordering, and surface orientation more faithfully than PixelGen.

\begin{table}[t]
\centering
\caption{\textbf{Partial-noise reconstruction shows stronger geometry preservation for PixelDense.} PixelGen and PixelDense reconstruct $\hat{\bx}_0$ from $\bx_t=(1-\tau)\bx_0+\tau\varepsilon$ at three noise levels. OneFormer PQ/mIoU are averaged over Swin-L and DiNAT-L COCO panoptic models; Depth cells report mean/median AbsRel after per-image affine fitting; normal cells report mean angular error in degrees. Pseudo references are obtained by applying the predictor to the original image after the shared resize and center crop. Higher is better for PQ and mIoU; lower is better for depth and normals.}
\label{tab:geometry_reconstruction}
\resizebox{\textwidth}{!}{%
\begin{tabular}{l l c | c c | c c | c c | c c}
\toprule
\multirow{2}{*}{Dataset} & \multirow{2}{*}{Reference} & \multirow{2}{*}{$\tau$} & \multicolumn{2}{c|}{OneFormer PQ $\uparrow$} & \multicolumn{2}{c|}{OneFormer mIoU $\uparrow$} & \multicolumn{2}{c|}{Depth AbsRel $\downarrow$} & \multicolumn{2}{c}{Normal err. ($^\circ$) $\downarrow$} \\
 & & & PixelGen & PixelDense & PixelGen & PixelDense & PixelGen & PixelDense & PixelGen & PixelDense \\
\midrule
\multirow{6}{*}{COCO val} & \multirow{3}{*}{Real panoptic} & 0.5 & 23.23 & \textbf{31.43} & 37.67 & \textbf{45.47} & -- & -- & -- & -- \\
 & & 0.7 & 15.70 & \textbf{21.31} & 30.64 & \textbf{36.59} & -- & -- & -- & -- \\
 & & 0.9 & 7.99 & \textbf{9.64} & 22.19 & \textbf{24.26} & -- & -- & -- & -- \\
\cmidrule(lr){2-11}
 & \multirow{3}{*}{Pseudo orig.} & 0.5 & 29.53 & \textbf{41.16} & 38.70 & \textbf{47.99} & 1.58/1.06 & \textbf{1.06/0.68} & 21.80 & \textbf{17.84} \\
 & & 0.7 & 18.93 & \textbf{26.51} & 30.73 & \textbf{37.32} & 2.09/1.54 & \textbf{1.63/1.12} & 26.90 & \textbf{23.48} \\
 & & 0.9 & 9.28 & \textbf{11.40} & 21.86 & \textbf{24.11} & 2.83/2.24 & \textbf{2.61/2.03} & 33.20 & \textbf{31.94} \\
\midrule
\multirow{3}{*}{Flickr30K} & \multirow{3}{*}{Pseudo orig.} & 0.5 & 19.66 & \textbf{30.09} & 19.51 & \textbf{28.73} & 1.61/1.14 & \textbf{1.03/0.68} & 22.45 & \textbf{18.25} \\
 & & 0.7 & 11.03 & \textbf{16.70} & 14.41 & \textbf{18.62} & 2.22/1.70 & \textbf{1.66/1.21} & 28.12 & \textbf{24.62} \\
 & & 0.9 & 4.19 & \textbf{5.90} & 9.47 & \textbf{10.94} & 3.01/2.49 & \textbf{2.77/2.31} & 34.97 & \textbf{33.88} \\
\bottomrule
\end{tabular}%
}
\end{table}

\noindent\textbf{REPA Teacher Composition Analysis.}
As shown in \Cref{tab:teacher_compose}, we compare four settings: adding one REPA teacher to the original PixelGen backbone, naively combining four REPA teachers, using factored REPA, and removing one REPA teacher at a time from PixelDense. 
From the results, each added dense REPA teacher improves GenEval Overall over DINOv2 alone. 
Depth Anything v2 and Metric3D v2 reach $0.8069$ and $0.8060$, respectively, exceeding SAM2 at $0.8020$. 
This agrees with iREPA's finding that spatial structure carries the alignment signal~\citep{singh2025matters}.
The naive four-teacher REPA sum performs worse than the best single dense REPA teacher. 
Thus, when semantic and geometric losses update the same diffusion-denoiser projection, adding more targets does not necessarily provide better supervision. 
Full PixelDense reaches $0.8093$, outperforming all single-teacher and unfactored variants. 
Using either stream alone gives a lower score, which indicates that semantic and geometric cues are complementary under the factored objective. 
The leave-one-out ablation further shows that each REPA teacher contributes to the final model.

\begin{table*}[t]
\small
\centering
\begin{minipage}[t]{0.48\linewidth}
    \centering
    \caption{\textbf{REPA Teacher Composition.} GenEval scores for teacher combinations. Baseline (DINOv2 only, PixelGen) is 0.7927. \textbf{DA2}: Depth Anything v2; \textbf{M3D}: Metric3D v2.}
    \label{tab:teacher_compose}
    \setlength{\tabcolsep}{3pt}
    \resizebox{\linewidth}{!}{%
    \begin{tabular}{lc@{\hspace{6pt}}lc}
    \toprule
    \textbf{REPA teacher} & \textbf{GenEval} & \textbf{REPA teacher} & \textbf{GenEval} \\
    \midrule
    \rowcolor[gray]{.95}
    \multicolumn{2}{l}{\textit{(a) Single + DINOv2}} & \multicolumn{2}{l}{\textit{(c) Factored streams}} \\
    $+$ SAM2 & 0.8020 & Semantic (D+S)         & 0.8022 \\
    $+$ DA2  & 0.8069 & Geometric (DA2+M3D)    & 0.7960 \\
    $+$ M3D  & 0.8060 & \textbf{Full PixelDense} & \textbf{0.8093} \\
    \midrule
    \rowcolor[gray]{.95}
    \multicolumn{2}{l}{\textit{(b) Unfactored sum}} & \multicolumn{2}{l}{\textit{(d) Leave-one-out}} \\
    Equal weight    & 0.8036 & w/o DINOv2 & 0.7940 \\
    L2-normalized   & 0.7970 & w/o SAM2   & 0.8027 \\
    Reduced weights & 0.8009 & w/o DA2    & 0.8025 \\
                    &        & w/o M3D    & 0.8037 \\
    \bottomrule
    \end{tabular}}
\end{minipage}
\hfill
\begin{minipage}[t]{0.48\linewidth}
    \centering
    \caption{\textbf{Ablation on PixelDense Design Choices.} (a) Bottleneck width, (b) Alignment block, (c) REPA loss weights, and (d) Orthogonality weight $\lambda_{\text{orth}}$. Bold indicates default.}
    \label{tab:ablations}
    \setlength{\tabcolsep}{3pt}
    \resizebox{\linewidth}{!}{%
    \begin{tabular}{lc@{\hspace{6pt}}lc}
    \toprule
    \textbf{Configuration} & \textbf{GenEval} & \textbf{Configuration} & \textbf{GenEval} \\
    \midrule
    \rowcolor[gray]{.95}
    \multicolumn{2}{l}{\textit{(a) Bottleneck (S/G)}} & \multicolumn{2}{l}{\textit{(c) REPA wts. (D/S/D/M)}} \\
    $512/512$                    & 0.8053          & $0.3/0.3/0.3/0.3$               & 0.8079 \\
    $\mathbf{768/1024}$ (def.)   & \textbf{0.8093} & $\mathbf{0.5/0.3/0.3/0.3}$ (def.) & \textbf{0.8093} \\
    $1024/1024$                  & 0.8065          & $0.5/0.5/0.5/0.5$               & 0.8081 \\
    \midrule
    \rowcolor[gray]{.95}
    \multicolumn{2}{l}{\textit{(b) Alignment block}} & \multicolumn{2}{l}{\textit{(d) $\lambda_{\text{orth}}$}} \\
    Block 6                 & 0.8075          & $0$                  & 0.8053 \\
    \textbf{Block 8} (def.) & \textbf{0.8093} & $0.001$              & 0.8077 \\
    Block 10                & 0.8007          & $\mathbf{0.01}$ (def.) & \textbf{0.8093} \\
                            &                 & $0.1$                & 0.8023 \\
    \bottomrule
    \end{tabular}}
\end{minipage}
\end{table*}

\subsection{Ablation Studies}
\label{sec:exp_ablations}

\Cref{tab:ablations} ablates four design choices in PixelDense: the per-stream projection bottleneck width, the alignment block, the per-teacher REPA loss weights, and the orthogonality regularizer. Each row varies one factor relative to the default of \Cref{tab:main_compare}.

\noindent\textbf{Projection Bottleneck Width.} The default widths match the native feature dimensions of the two teacher groups. Shrinking both streams to a smaller shared width removes capacity needed by the geometric teachers, while equalizing them at the larger width wastes semantic-stream parameters. Both variants underperform the default PixelDense configuration.

\noindent\textbf{Alignment Block.} The mid-network tap is strongest, consistent with REPA~\citep{yu2025representation}. Moving the tap one block earlier causes a small drop, while moving it later causes a larger drop. A mid-network position leaves enough upstream computation to form teacher-compatible features and enough downstream capacity to decode them into pixels.

\noindent\textbf{Alignment-Loss Weights.} Replacing the default weighting with uniform weights yields scores close to the default, showing that PixelDense is robust to moderate weight changes. Retaining the slightly larger DINOv2 weight from the original pretrained objective still gives the best aggregate.

\noindent\textbf{Orthogonality Regularization.} Removing the penalty or making it $10\times$ weaker lowers GenEval Overall, and a $10\times$ stronger penalty over-constrains the projections. The default $\lambda_\text{orth}=0.01$ separates the two streams enough to lift compositional alignment without sacrificing capacity.

\begin{wrapfigure}{r}{0.45\textwidth}
    \centering
    \includegraphics[width=0.45\textwidth]{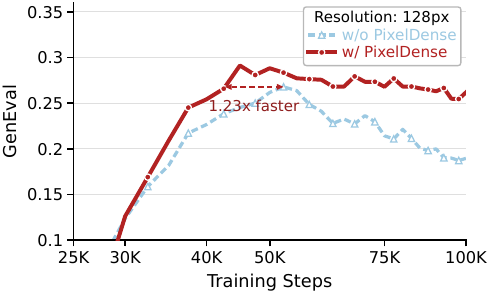}
    \caption{\textbf{PixelDense converges faster on GenEval from random initialization.}} %
    \label{fig:convergence_scratch}
\end{wrapfigure}

\subsection{Training from Scratch}
\label{sec:exp_scratch}
Our main runs fine-tune from a pretrained model that already carries DINOv2 alignment. 
To test whether dense teachers also help under scratch training, we train a pixel diffusion model from random initialization at $128 \times 128$ resolution on BLIP3-o-60K~\citep{chen2025blip3}, with and without PixelDense under matched optimizer, batch size, and data ordering. \Cref{fig:convergence_scratch} reports GenEval Overall against optimizer step. PixelDense reaches the peak score of the baseline roughly $1.23\times$ faster and stays above it through the rest of the $100$K-step horizon, while the baseline drifts downward after its peak. The result is consistent with \Cref{tab:main_compare}: dense alignment helps the denoiser form teacher-compatible features earlier, yielding both faster convergence and a higher final score.

\subsection{Transfer to DeCo}
\label{sec:exp_deco}
To test whether the recipe ports beyond PixelGen, we apply the same setting to DeCo without retuning. 
As shown in \Cref{tab:deco_transfer}, PixelDense improves DeCo on both GenEval Overall and DPG-Bench over the matched DeCo fine-tune. 
The gain is smaller than on PixelGen likely because DeCo starts from a stronger model, but the direction is consistent, indicating that the recipe ports to a different pixel-space backbone.

\begin{table}[t]
\centering
\caption{\textbf{The PixelDense recipe transfers to the DeCo backbone without retuning.} The teachers, losses, hyperparameters, and checkpoint-selection rule used on PixelGen-XXL are applied unchanged to DeCo~\citep{ma2025deco} at $512 \times 512$. Higher is better for every metric, and bold marks PixelDense entries that improve over the matched DeCo fine-tune.}
\label{tab:deco_transfer}
\resizebox{0.9\textwidth}{!}{%
\begin{tabular}{l | c c c c c c c | c}
\toprule
\multirow{2}{*}{Model} & \multicolumn{7}{c|}{GenEval $\uparrow$} & \multirow{2}{*}{DPG $\uparrow$} \\
 & Sing. Obj. & Two Obj. & Count. & Colors & Attr. & Pos. & Overall & \\
\midrule
DeCo T2I, released checkpoint~\citep{ma2025deco} & 100.00 & 92.00 & 72.00 & 91.00 & 79.00 & 80.00 & 0.8600 & 81.4 \\
DeCo + fine-tune & 99.38 & \textbf{94.44} & 73.75 & \textbf{93.62} & 79.00 & 77.00 & 0.8620 & 81.4 \\
\textbf{PixelDense} & \textbf{100.00} & 93.43 & \textbf{74.38} & \textbf{93.62} & \textbf{80.00} & \textbf{80.00} & \textbf{0.8690} & \textbf{81.8} \\
\bottomrule
\end{tabular}%
}
\end{table}

\subsection{Transfer to Image Editing}
\label{sec:exp_edit}
\Cref{tab:geometry_reconstruction} shows that PixelDense keeps more of the input structure when it reconstructs a noised image. Editing needs the same ability, since the model should change the edited region and keep the rest. We run SDEdit~\citep{meng2022sdedit} on PIE-Bench~\citep{ju2024pnp}: the source image is noised to level $\tau$ as in the reconstruction probe and then denoised with the target prompt, for seven levels $\tau \in \{0.3, 0.4, \dots, 0.9\}$. Both models use the same source images, target prompts, noise, sampler, and number of steps, so only the checkpoint differs. We measure background preservation outside the PIE-Bench edit mask with LPIPS, PSNR, and SSIM. For structure, we use DINOv2 patch self-similarity, which compares the patch-to-patch similarity maps of the edited and source images, on the full image and on the background. We also report FID against the source images. As shown in \Cref{tab:pie_edit}, PixelDense is better on all six metrics at every noise level. The FID and DINOv2 gaps shrink at high $\tau$, where little of the source is left for either model.

\begin{table}[t]
\centering
\caption{\textbf{PixelDense keeps more of the source image in SDEdit editing on PIE-Bench.} Only the checkpoint differs between the two models. Background metrics are computed outside the edit mask, and FID is computed against the source images. Bold marks PixelDense entries that improve over PixelGen.}
\label{tab:pie_edit}
\begin{tabular}{l l | c c c c c c c}
\toprule
\multirow{2}{*}{Metric} & \multirow{2}{*}{Model} & \multicolumn{7}{c}{Noise level $\tau$} \\
 & & 0.3 & 0.4 & 0.5 & 0.6 & 0.7 & 0.8 & 0.9 \\
\midrule
\multicolumn{9}{l}{\emph{Background, outside the edit mask}} \\
\multirow{2}{*}{LPIPS $\downarrow$} & PixelGen & 0.22 & 0.24 & 0.26 & 0.28 & 0.30 & 0.32 & 0.35 \\
 & PixelDense & \textbf{0.19} & \textbf{0.21} & \textbf{0.23} & \textbf{0.24} & \textbf{0.26} & \textbf{0.29} & \textbf{0.32} \\
\multirow{2}{*}{PSNR (dB) $\uparrow$} & PixelGen & 26.07 & 24.72 & 23.27 & 21.76 & 20.17 & 18.19 & 15.57 \\
 & PixelDense & \textbf{28.00} & \textbf{26.74} & \textbf{25.39} & \textbf{23.92} & \textbf{22.30} & \textbf{20.23} & \textbf{17.10} \\
\multirow{2}{*}{SSIM $\uparrow$} & PixelGen & 0.80 & 0.77 & 0.74 & 0.71 & 0.69 & 0.66 & 0.62 \\
 & PixelDense & \textbf{0.84} & \textbf{0.82} & \textbf{0.80} & \textbf{0.77} & \textbf{0.74} & \textbf{0.71} & \textbf{0.66} \\
\midrule
\multicolumn{9}{l}{\emph{DINOv2 self-similarity to the source}} \\
\multirow{2}{*}{Full image $\uparrow$} & PixelGen & 0.65 & 0.60 & 0.55 & 0.51 & 0.47 & 0.43 & 0.38 \\
 & PixelDense & \textbf{0.72} & \textbf{0.67} & \textbf{0.62} & \textbf{0.58} & \textbf{0.53} & \textbf{0.48} & \textbf{0.41} \\
\multirow{2}{*}{Background $\uparrow$} & PixelGen & 0.80 & 0.77 & 0.74 & 0.72 & 0.69 & 0.67 & 0.63 \\
 & PixelDense & \textbf{0.84} & \textbf{0.81} & \textbf{0.78} & \textbf{0.76} & \textbf{0.73} & \textbf{0.70} & \textbf{0.65} \\
\midrule
\multirow{2}{*}{FID $\downarrow$} & PixelGen & 53.99 & 59.93 & 65.28 & 69.21 & 71.11 & 74.19 & 77.54 \\
 & PixelDense & \textbf{41.85} & \textbf{48.99} & \textbf{54.72} & \textbf{60.72} & \textbf{65.17} & \textbf{69.26} & \textbf{73.51} \\
\bottomrule
\end{tabular}
\end{table}

\subsection{Qualitative Results}
\label{sec:exp_qual}

\begin{figure}[t]
    \centering
    \includegraphics[width=0.95\textwidth]{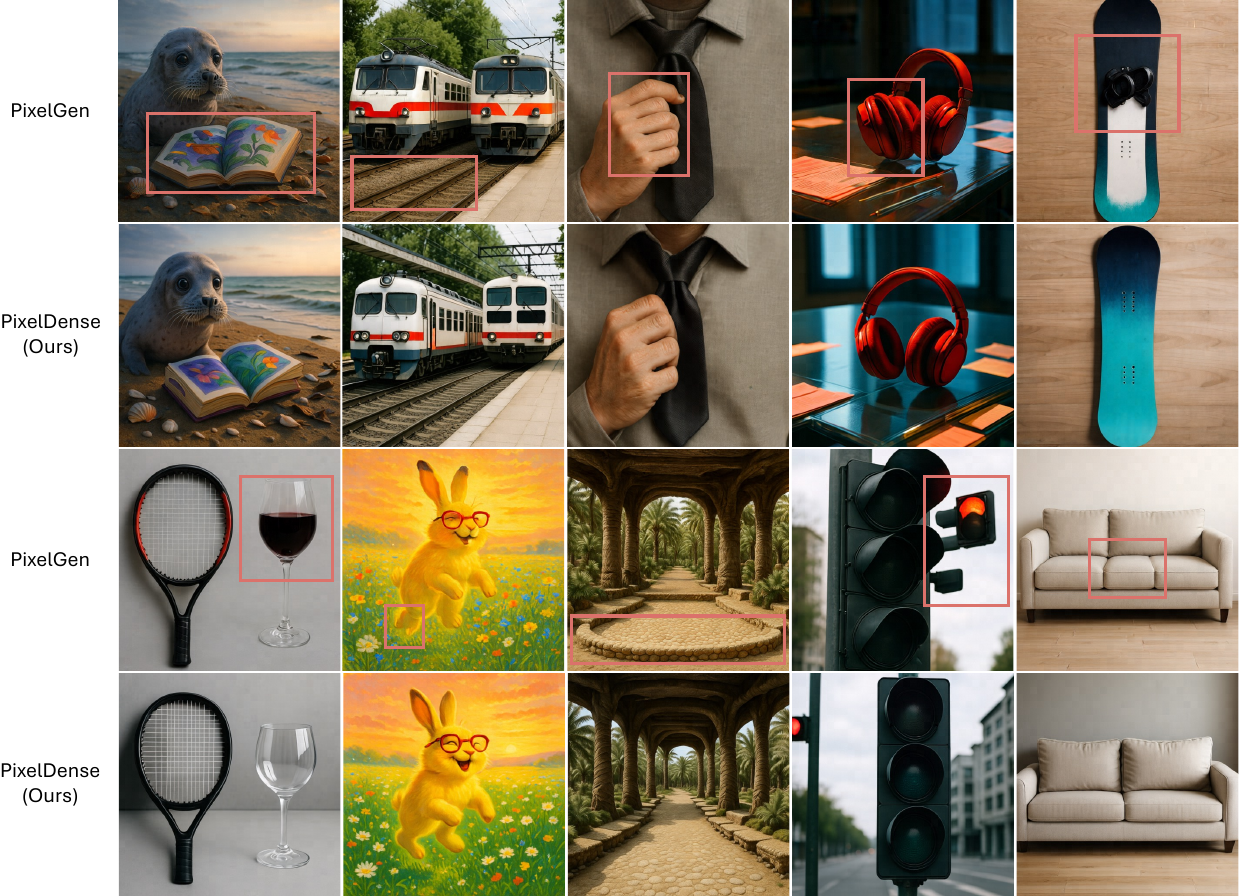}

    \caption{\textbf{Qualitative comparison between PixelGen and our PixelDense on text-to-image generation.} PixelDense better preserves object counts, spatial relations, color binding, and scene geometry than PixelGen, yielding stronger prompt alignment and visual coherence.}
    \label{fig:qualitative_compare}
\end{figure}

\Cref{fig:qualitative_compare} compares PixelGen and PixelDense on matched prompts and seeds, with the failure modes selected to track the quantitative gains: object count, relative position, color binding, object boundaries, and perspective. All samples use the same $25$-step sampler as the quantitative evaluation, and the dense teachers are absent at inference.

\section{Limitation and Conclusion}
\label{sec:conclusion}

This paper introduces PixelDense, a dense-perception representation-alignment framework for pixel-space diffusion. 
Our approach routes semantic and geometric teachers through separate streams, using weight-space orthogonality to prevent collapse into a single denoiser subspace.
Experiments on PixelGen and DeCo show improved compositional alignment at no inference-time cost, while reconstruction probes show stronger preservation of segmentation, depth, and surface-normal structure. 
Image editing typically requires modifying content while maintaining source image identity, layout, and 3D structure.
With SDEdit on PIE-Bench, PixelDense keeps more of the source background and layout at every edit strength, so it is a promising approach for scenarios that demand strict structure preservation.
Our editing tests cover only SDEdit, and other editing methods, such as inversion-based and instruction-based editing, are left to future work.
We hope PixelDense encourages using dense perception models as practical training-time supervision for image generators.

\begin{ack}
  This work is supported in part by the U.S.~Office of Naval Research Award under Grant Number N00014-24-1-2668, the National Science Foundation under Grants IIS-2316306 and CNS-2330215, the National Institutes of Health (NIH) under Grant R01EB293388, Merck BARDS Academic Collaboration Grant, and gifts from Adobe Research.
\end{ack}

\bibliographystyle{plainnat}
\bibliography{references}

\appendix
\clearpage

\appendix
\section*{Appendix}

\section{Implementation Details}
\label{sec:appendix_impl}
\subsection{Pixel-Space Backbones}
\label{sec:appendix_impl_backbones}

The primary backbone is PixelGen-XXL~\citep{ma2026pixelgen}, a $1.1$B-parameter pixel-space diffusion transformer with $16$ visual blocks. 
Inputs are $512 \times 512$ RGB images split into $32 \times 32$ patches, giving $N = 256$ visual tokens of hidden width $d_h = 1536$. 
The text branch uses a frozen Qwen3 text encoder~\citep{yang2025qwen3} with cross-attention into the visual stream. 
The transfer backbone is DeCo-T2I~\citep{ma2025deco}, a $1.1$B pixel diffusion transformer with the same patch grid. 
Both backbones are pretrained with single-teacher DINOv2~\citep{oquab2023dinov2} REPA~\citep{yu2025representation} and continue to receive that signal during PixelDense fine-tuning. The deployed sampler reads only the three RGB channels, so PixelDense adds no parameters at inference.

\subsection{Frozen Dense Teachers}
\label{sec:appendix_impl_teachers}

The four teachers and the block used for alignment are summarized in \Cref{tab:teacher_specs}. 
Each teacher consumes the clean image $\br$ scaled to $[0,1]$ at $512 \times 512$, and emits a token grid that is bilinearly resampled to the denoiser's $16 \times 16$ patch lattice and channel-wise L2 normalized before the cosine alignment of \Cref{eq:single_repa,eq:factored_teacher_loss}. 
Teacher weights remain frozen, no augmentation is applied at the teacher input, and gradients flow only through the denoiser, the projection streams, and the per-teacher heads.

\begin{table}[h]
\centering
\caption{\textbf{Frozen dense-prediction teachers used by PixelDense.} All teachers operate on the clean image $\br$ at $512 \times 512$, are frozen during training, and are dropped at inference. Feature dimension is the per-token width before normalization. Group is the semantic-versus-geometric assignment used by \Cref{sec:method_pixeldense}.}
\label{tab:teacher_specs}
\setlength{\tabcolsep}{4pt}
\begin{tabular}{l l l c l}
\toprule
Teacher & Backbone & Tap & Feature dim. & Group \\
\midrule
DINOv2~\citep{oquab2023dinov2}                   & ViT-B/14 patch tokens          & last layer  & $768$  & semantic \\
SAM2~\citep{ravi2024sam}                         & Hiera-L image encoder          & last layer  & $256$  & semantic \\
Depth Anything v2~\citep{yang2024depthv2}        & ViT-L encoder                  & layer $17$  & $1024$ & geometric \\
Metric3D v2~\citep{hu2024metric3d}               & ViT-L encoder                  & layer $17$  & $1024$ & geometric \\
\bottomrule
\end{tabular}
\end{table}

\subsection{Projection Streams and Teacher Heads}
\label{sec:appendix_impl_streams}

The semantic stream $\pi_s$ and the geometric stream $\pi_g$ both read the block-$8$ token state $\bh_\ell \in \R^{N \times d_h}$ of the denoiser. Each stream is a two-layer MLP of the form $W^{(2)} \sigma(\mathrm{LN}(W \bh_{\ell,n}))$, with SiLU nonlinearity $\sigma$ and channel-wise LayerNorm. The semantic stream uses $W_\text{sem} \in \R^{768 \times d_h}$ followed by $W_\text{sem}^{(2)} \in \R^{768 \times 768}$, and the geometric stream uses $W_\text{geo} \in \R^{1024 \times d_h}$ followed by $W_\text{geo}^{(2)} \in \R^{1024 \times 1024}$. The four teacher heads $g_\text{DINO}, g_\text{SAM}, g_\text{DA2}, g_\text{M3D}$ are single linear maps from the corresponding stream width to the teacher feature dimension in \Cref{tab:teacher_specs}. The orthogonality penalty of \Cref{eq:orth} acts on the row-normalized matrices $\bar W_\text{sem}$ and $\bar W_\text{geo}$ and adds no inference-time cost.

\subsection{Training Recipe}
\label{sec:appendix_impl_train}

Fine-tuning runs use $2 \times$ NVIDIA H200 GPUs with a per-GPU batch size of $16$ and $8$ gradient-accumulation steps, giving an effective batch size of $256$. The optimizer is AdamW with $\beta_1 = 0.9$, $\beta_2 = 0.999$, weight decay $0.01$, learning rate $1\text{e-}5$, and a constant schedule after a $200$-step linear warmup. Each fine-tune runs for $10{,}000$ optimizer steps, which is the budget reported in \Cref{tab:main_compare,tab:teacher_compose,tab:ablations,tab:deco_transfer}. The full objective of \Cref{eq:total} combines the flow-matching loss $\mathcal{L}_\text{FM}$, the four cosine alignment losses with the weights in \Cref{eq:align}, the orthogonality penalty with $\lambda_\text{orth} = 0.01$, and the auxiliary LPIPS and Perceptual-DINO losses inherited from PixelGen. We use bf16 mixed precision, gradient norm clipping at $1.0$, and exponential moving average over denoiser parameters with decay $0.9999$. The same recipe transfers to DeCo without retuning. The from-scratch run of \Cref{fig:convergence_scratch} trains a pixel diffusion model from random initialization at $128 \times 128$ on BLIP3-o-60K~\citep{chen2025blip3} with the same effective batch size and optimizer, for $100{,}000$ steps with and without PixelDense.

\subsection{Sampling}
\label{sec:appendix_impl_sample}

At inference, the dense teachers and the alignment-side parameters $\Phi$ are dropped. 
We use the Adam LM solver with $25$ steps and classifier-free guidance scale $4.0$. Generation resolution is $512 \times 512$ for PixelGen-XXL and DeCo, and $128 \times 128$ for the from-scratch run. 
Reported numbers are seed-matched per evaluation, so PixelGen, the matched fine-tune, and PixelDense receive the same prompts and sampler.

\subsection{Datasets}
\label{sec:appendix_impl_data}

Fine-tuning uses BLIP3-o-60K~\citep{chen2025blip3}, the third-stage dataset of PixelGen training, with no extra filtering. The from-scratch run uses the same dataset at $128 \times 128$. Evaluation uses GenEval~\citep{ghosh2023geneval}, DPG-Bench~\citep{hu2024ella}, and HPS v2.1~\citep{wu2023human} for text-to-image quality, COCO 2017~\citep{lin2014microsoft} val and Flickr30K~\citep{plummer2015flickr30k} test for partial-noise reconstruction. 
All evaluation images are pre-resized with bilinear interpolation to $512 \times 512$ followed by a center crop, and GT panoptic masks are nearest resampled with the same crop so that pixel indices align across reconstruction, prediction, and reference. We did not retrain or fine-tune any teacher, probe, or evaluation model.

\section{Evaluation Protocols}
\label{sec:appendix_eval}

\subsection{Text-to-Image Benchmarks}
\label{sec:appendix_eval_t2i}

GenEval~\citep{ghosh2023geneval} is run with the released $553$ prompt set across the six standard categories of single-object, two-object, counting, colors, attribute binding, and position. 
Score with the official GenEval scorer, and report Overall together with the six per-category scores. DPG-Bench~\citep{hu2024ella} is run with its official prompt set, four images per prompt, and the released grader. HPS v2.1~\citep{wu2023human} is run on the four $800$-prompt style splits of Anime, Concept-art, Paintings, and Photo, with one image per prompt. The per-style breakdown for our PixelGen-XXL fine-tunes is given in \Cref{tab:hps_style}.

\subsection{Partial-Noise Reconstruction Probes}
\label{sec:appendix_eval_probes}

The probe of \Cref{tab:geometry_reconstruction} corrupts an input image $\bx_0$ by $\bx_t = (1 - \tau) \bx_0 + \tau \varepsilon$ with $\varepsilon \sim \mathcal{N}(0, I)$, feeds $\bx_t$ together with a caption to the model, and reads the predicted clean image $\hat{\bx}_0$. Captions are the original COCO caption with the lowest sentence index for COCO, and the first non-empty Flickr30K caption otherwise. For each $\tau \in \{0.5, 0.7, 0.9\}$ we use a single fixed Gaussian sample $\varepsilon$ shared across PixelGen and PixelDense, so the two models reconstruct from identical noisy inputs.

\paragraph{Panoptic probes.}
The same OneFormer~\citep{jain2023oneformer} model with two backbones, Swin-L~\citep{liu2021swin} and DiNAT-L~\citep{hassani2022dilated} pretrained on COCO panoptic for $100$ epochs, is applied to the reconstruction $\hat{\bx}_0$ and the reference, both at $512 \times 512$. Real-COCO references are the public panoptic GT after nearest resize and the same center crop. Pseudo references are the OneFormer prediction on the resize-cropped original. We score with PQ~\citep{kirillov2019panoptic} and mIoU under the COCO panoptic vocabulary, and average across Swin-L and DiNAT-L for the cells of \Cref{tab:geometry_reconstruction}. Per-backbone numbers and PQ$_\text{th}$/PQ$_\text{st}$ breakdowns are in \Cref{tab:appendix_oneformer_swin,tab:appendix_oneformer_dinat}.

\paragraph{Depth probe.}
We run Marigold v1.1 depth~\citep{ke2024repurposing} on the reconstruction $\hat{\bx}_0$ and on the resize-cropped original. The original prediction is treated as a pseudo-reference because COCO and Flickr30K do not ship dense metric depth. 
For each image we solve the closed-form least-squares scale and shift $(a, b)$ that align $a \cdot d_\text{recon} + b$ to $d_\text{orig}$, exclude pixels with $d_\text{orig} \leq 10^{-3}$, and report mean and median of pixelwise AbsRel, $\lvert a \cdot d_\text{recon} + b - d_\text{orig} \rvert / d_\text{orig}$.

\paragraph{Surface-normal probe.}
We run Marigold v1.1 normals on the reconstruction and the resize-cropped original, defensively re-normalize both to unit length, and report the mean pixelwise angular error $\arccos(\langle \hat n_\text{recon}, \hat n_\text{orig}\rangle)$ in degrees. Per-image median and mean-of-per-image-medians statistics are in \Cref{tab:appendix_marigold_normals}.

\paragraph{Datasets and counts.}
COCO val 2017 contributes $5{,}000$ images to every probe, real-GT and pseudo-reference. Flickr30K test contributes $1{,}000$ images filtered to non-empty captions, pseudo reference only. The same image set is used for PixelGen and PixelDense at every $\tau$, so the comparisons in \Cref{tab:geometry_reconstruction} are paired across models.

\section{Additional Results}
\label{sec:appendix_results}

\subsection{Semantic versus Geometric Teacher Features}
\label{sec:appendix_teacher_viz}
\begin{figure}[h]
    \centering
    \includegraphics[width=0.85\textwidth]{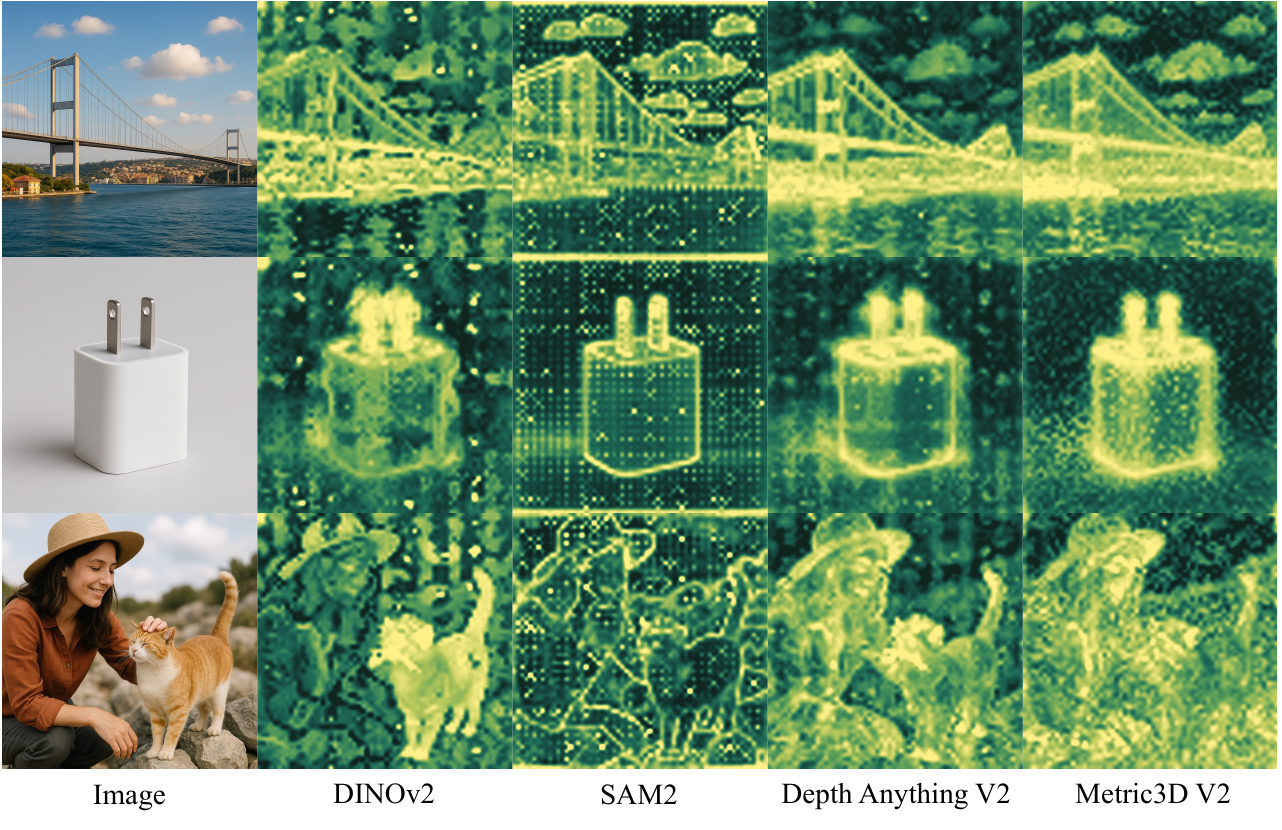}
    \vspace{-2mm}
    \caption{\textbf{Semantic and geometric teachers light up complementary spatial structure on the same image.} The leftmost column shows the RGB input, and the next four columns show the per-patch RMS magnitude of spatially normalized features from DINOv2, SAM2, Depth Anything v2, and Metric3D v2, with yellow marking high response. DINOv2 fires on object regions and SAM2 fires on instance boundaries, while Depth Anything v2 traces depth-ordered silhouettes and Metric3D v2 traces surface-normal layout.}
    \label{fig:teacher_feature_viz}
    \vspace{-2mm}
\end{figure}

\Cref{fig:teacher_feature_viz} visualizes the per-patch responses of the four dense teachers on the same image. Semantic teachers fire on object regions and instance boundaries, while geometric teachers trace depth-ordered silhouettes and surface-normal layout. The split motivates the semantic-versus-geometric factorization of \Cref{sec:method_pixeldense}.

\subsection{Training-Dynamics Diagnostic for Unfactored versus Factored Alignment}
\label{sec:appendix_dynamics}

This subsection looks at the training trajectories. We log the four per-teacher cosine alignment losses every $50$ optimizer steps and read the four curves jointly across runs from \Cref{tab:teacher_compose}, the naive equal-weight sum and our default PixelDense configuration. The diagnostic examines three properties simultaneously, namely whether the four teachers co-improve over the full training, whether they move in lockstep at the optimizer-step scale, and whether each teacher's signal is being absorbed at a similar rate.

\paragraph{Setup.}
Both runs share the recipe of \Cref{sec:exp_setup}. We log every $50$ optimizer steps from step $49$ to step $9{,}999$, giving $200$ points per run for each of $\mathcal{L}_\text{DINO}$, $\mathcal{L}_\text{SAM}$, $\mathcal{L}_\text{DA2}$, and $\mathcal{L}_\text{M3D}$. We compute three diagnostics on this $200 \times 4$ matrix per run. Pearson correlation on the centered curves measures whether the four losses co-improve over the full schedule. Cosine similarity of step-to-step loss deltas measures whether the four losses move together or apart at the optimizer-step scale. Per-teacher relative loss reduction, taken between the first ten and last ten logged points, measures how much of each teacher's signal the denoiser has absorbed.

\begin{table}[t]
\centering
\caption{\textbf{PixelDense preserves long-term co-improvement across teachers while breaking the short-term lockstep that the naive sum imposes.} Run-level coupling and balance diagnostics on the four cosine alignment losses logged every $50$ optimizer steps for $10{,}000$ steps, giving $200$ points per run. Pearson is the off-diagonal mean of the $4 \times 4$ cross-loss correlation on those $200$ points. $\Delta_\text{cos}$ is the off-diagonal mean of the cosine on the $199$ consecutive step-to-step loss deltas. Pearson $-\, \Delta_\text{cos}$ isolates long-term co-improvement from short-term lockstep. Mean reduction is the average per-teacher relative loss reduction between the first and last ten logged points. Sem.\ and Geo.\ reduction restrict that average to the two semantic teachers and the two geometric teachers, and Sem./Geo.\ gap is their absolute difference. Reduction CV is the across-teacher coefficient of variation of the four reductions. Higher Pearson, higher Pearson $-\, \Delta_\text{cos}$, higher mean reduction, lower Sem./Geo.\ gap, and lower CV are better.}
\label{tab:repa_dynamics_summary}
\resizebox{\textwidth}{!}{%
\begin{tabular}{l | c c c | c c c c c}
\toprule
Run & Pearson $\uparrow$ & $\Delta_\text{cos}$ $\downarrow$ & Pearson $-\, \Delta_\text{cos}$ $\uparrow$ & Mean red.\ $\uparrow$ & Sem.\ red. & Geo.\ red. & Sem./Geo.\ gap $\downarrow$ & Reduction CV $\downarrow$ \\
\midrule
Naive sum, equal weight                             & 0.9057 & 0.7430 & 0.1627 & 0.7213 & 0.7573 & 0.6854 & 0.0719 & 0.0973 \\
\textbf{PixelDense}                         & \textbf{0.9083} & \textbf{0.5966} & \textbf{0.3116} & \textbf{0.7340} & 0.7337 & 0.7343 & \textbf{0.0006} & \textbf{0.0629} \\
\bottomrule
\end{tabular}%
}
\end{table}

\begin{table}[t]
\centering
\caption{\textbf{Step-to-step loss-delta cosine matrices on the four cosine alignment losses, naive sum versus PixelDense.} Each off-diagonal entry is the cosine of step-to-step deltas of the two named per-teacher losses, computed on the $199$ consecutive deltas of the $200$ logged points per run. The top sub-block reports the naive equal-weight sum, and the bottom sub-block reports PixelDense. The within-geometric Depth Anything v2 against Metric3D pair stays near $0.95$ under both objectives, reflecting the shared 3D structure of the two geometric teachers. The other five off-diagonal entries collapse only under PixelDense, with the largest drop on the within-semantic DINOv2 against SAM2 pair.}
\label{tab:repa_dynamics_delta}
\begin{tabular}{l | c c c c}
\toprule
 & DINOv2 & SAM2 & DA2 & M3D \\
\midrule
\multicolumn{5}{l}{\emph{Naive sum, equal weight, off-diagonal mean $0.7430$}} \\
DINOv2  & 1.0000 & 0.7287 & 0.7120 & 0.6950 \\
SAM2    & 0.7287 & 1.0000 & 0.6799 & 0.6671 \\
DA2     & 0.7120 & 0.6799 & 1.0000 & 0.9756 \\
M3D     & 0.6950 & 0.6671 & 0.9756 & 1.0000 \\
\midrule
\multicolumn{5}{l}{\emph{PixelDense, off-diagonal mean $0.5966$}} \\
DINOv2  & 1.0000 & \textbf{0.3567} & \textbf{0.5323} & \textbf{0.4969} \\
SAM2    & \textbf{0.3567} & 1.0000 & \textbf{0.6618} & \textbf{0.5843} \\
DA2     & \textbf{0.5323} & \textbf{0.6618} & 1.0000 & 0.9477 \\
M3D     & \textbf{0.4969} & \textbf{0.5843} & 0.9477 & 1.0000 \\
\bottomrule
\end{tabular}
\end{table}

\paragraph{Long-term co-improvement is preserved.}
As show in \Cref{tab:repa_dynamics_summary}, both runs have a high off-diagonal mean Pearson correlation across the four loss curves, $0.9057$ for the naive sum and $0.9083$ for PixelDense. The minimum cross-teacher Pearson is also high in both runs, $0.8079$ for naive and $0.8366$ for factored. The factoring does not break the basic property that all four cosine alignment losses descend together over the $10{,}000$-step horizon. The long-term coupling is marginally tighter under PixelDense.

\paragraph{Short-term lockstep weakens where teacher targets are not intrinsically correlated.}
The picture inverts at the optimizer-step scale. The off-diagonal mean of the step-delta cosine drops from $0.7430$ under the naive sum to $0.5966$ under PixelDense. \Cref{tab:repa_dynamics_delta} reports the full $4 \times 4$ matrices. The drop is concentrated on the five pairs that span teacher targets with weaker intrinsic correlation, the four cross-stream pairs and the within-semantic DINOv2-versus-SAM2 pair. The Depth Anything v2-versus-Metric3D pair stays at $0.9756$ under naive and $0.9477$ under PixelDense, which reflects the genuinely correlated 3D structure of the two geometric teachers and is the only pair that remains strongly coupled under both objectives. The DINOv2-versus-SAM2 step-delta cosine falls from $0.7287$ to $0.3567$ even though both losses share the semantic projector $\pi_s$, which says the per-teacher heads $g_\text{DINO}$ and $g_\text{SAM}$ are doing real decoupling work on top of the stream split. Read with the previous paragraph, this says that PixelDense preserves long-term co-progress while breaking the short-term lockstep that an unfactored summed objective imposes on the shared denoiser feature. The naive sum imposes that lockstep on teacher pairs whose targets are not intrinsically correlated. The combined effect is captured by the Pearson $-\, \Delta_\text{cos}$ column of \Cref{tab:repa_dynamics_summary}, which rises from $0.1627$ to $0.3116$.

\paragraph{Per-teacher progress is more balanced and the mean rises.}
\Cref{tab:repa_dynamics_progress} reports per-teacher relative reduction. The naive run reduces $\mathcal{L}_\text{DINO}$ by $83.10\%$ and $\mathcal{L}_\text{M3D}$ by only $64.26\%$. The $19$-point gap matches the picture of negative transfer between teacher losses sharing one denoiser feature, the summed update mostly follows the dominant gradient and leaves slower teachers under-served. Under PixelDense the DINOv2 reduction settles at $78.49\%$, the Depth Anything v2 reduction rises from $72.81\%$ to $77.49\%$, and the Metric3D normal reduction rises from $64.26\%$ to $69.37\%$. Mean reduction across the four teachers rises from $0.7213$ to $0.7340$. The semantic-versus-geometric reduction gap drops from $0.0719$ to $0.0006$, and the across-teacher coefficient of variation drops from $0.0973$ to $0.0629$. Each loss is being absorbed at a rate more uniform across the bank, with no teacher under-served, and the higher mean reduction is achieved without requiring more than a $4.6$-point relative drop on the dominant DINOv2 term.

\begin{table}[t]
\centering
\caption{\textbf{Per-teacher relative loss reduction over $10{,}000$ training steps, naive sum versus PixelDense.} The reduction is computed between the first ten and last ten logged points of the $200$-point training log per run. Higher values indicate that more of a teacher's signal has been absorbed. Under the naive sum, Metric3D is the slowest teacher in the bank at $64.26\%$ and DINOv2 is the fastest at $83.10\%$. PixelDense narrows the spread to within $10$ points across the four teachers and raises the mean from $72.13\%$ to $73.40\%$.}
\label{tab:repa_dynamics_progress}
\begin{tabular}{l | c c c c | c}
\toprule
Run & DINOv2 & SAM2 & DA2 & M3D & Mean \\
\midrule
Naive sum, equal weight                             & \textbf{83.10\%} & \textbf{68.35\%} & 72.81\% & 64.26\% & 72.13\% \\
\textbf{PixelDense}                         & 78.49\% & 68.26\% & \textbf{77.49\%} & \textbf{69.37\%} & \textbf{73.40\%} \\
\bottomrule
\end{tabular}
\end{table}

\begin{table}[t]
\centering
\caption{\textbf{OneFormer Swin-L panoptic scores per dataset, reference type, and noise fraction.} Reconstructions of $\hat{\bx}_0$ are scored against the indicated reference at $512 \times 512$. PQ is reported on All, Things, and Stuff. Higher is better, and bold marks the PixelDense column.}
\label{tab:appendix_oneformer_swin}
\resizebox{\textwidth}{!}{%
\begin{tabular}{l l c | c c | c c | c c | c c}
\toprule
\multirow{2}{*}{Dataset} & \multirow{2}{*}{Reference} & \multirow{2}{*}{$\tau$} & \multicolumn{2}{c|}{PQ$_\text{All}$ $\uparrow$} & \multicolumn{2}{c|}{PQ$_\text{Th}$ $\uparrow$} & \multicolumn{2}{c|}{PQ$_\text{St}$ $\uparrow$} & \multicolumn{2}{c}{mIoU $\uparrow$} \\
 & & & PixelGen & PixelDense & PixelGen & PixelDense & PixelGen & PixelDense & PixelGen & PixelDense \\
\midrule
\multirow{6}{*}{COCO val} & \multirow{3}{*}{Real panoptic} & 0.5 & 23.25 & \textbf{31.54} & 26.32 & \textbf{35.96} & 18.62 & \textbf{24.86} & 37.53 & \textbf{45.36} \\
 & & 0.7 & 15.69 & \textbf{21.31} & 17.16 & \textbf{23.93} & 13.48 & \textbf{17.37} & 30.61 & \textbf{36.49} \\
 & & 0.9 & 8.02 & \textbf{9.72} & 7.92 & \textbf{9.96} & 8.17 & \textbf{9.36} & 22.16 & \textbf{24.32} \\
\cmidrule(lr){2-11}
 & \multirow{3}{*}{Pseudo orig.} & 0.5 & 29.72 & \textbf{41.33} & 31.90 & \textbf{44.39} & 26.43 & \textbf{36.71} & 38.62 & \textbf{47.79} \\
 & & 0.7 & 19.01 & \textbf{26.65} & 19.95 & \textbf{28.14} & 17.59 & \textbf{24.39} & 30.76 & \textbf{37.38} \\
 & & 0.9 & 9.38 & \textbf{11.56} & 8.87 & \textbf{11.17} & 10.15 & \textbf{12.15} & 21.75 & \textbf{24.13} \\
\midrule
\multirow{3}{*}{Flickr30K} & \multirow{3}{*}{Pseudo orig.} & 0.5 & 19.35 & \textbf{29.19} & 16.54 & \textbf{25.50} & 23.29 & \textbf{34.35} & 19.07 & \textbf{28.03} \\
 & & 0.7 & 11.20 & \textbf{16.71} & 7.71 & \textbf{12.95} & 16.08 & \textbf{21.95} & 14.26 & \textbf{18.22} \\
 & & 0.9 & 4.33 & \textbf{6.11} & 2.26 & \textbf{4.09} & 7.23 & \textbf{8.92} & 9.30 & \textbf{10.79} \\
\bottomrule
\end{tabular}%
}
\end{table}

\begin{table}[t]
\centering
\caption{\textbf{OneFormer DiNAT-L panoptic scores per dataset, reference type, and noise fraction.} Same protocol as \Cref{tab:appendix_oneformer_swin}, with DiNAT-L replacing Swin-L. The two backbones agree to within $0.005$ on PQ and mIoU per cell.}
\label{tab:appendix_oneformer_dinat}
\resizebox{\textwidth}{!}{%
\begin{tabular}{l l c | c c | c c | c c | c c}
\toprule
\multirow{2}{*}{Dataset} & \multirow{2}{*}{Reference} & \multirow{2}{*}{$\tau$} & \multicolumn{2}{c|}{PQ$_\text{All}$ $\uparrow$} & \multicolumn{2}{c|}{PQ$_\text{Th}$ $\uparrow$} & \multicolumn{2}{c|}{PQ$_\text{St}$ $\uparrow$} & \multicolumn{2}{c}{mIoU $\uparrow$} \\
 & & & PixelGen & PixelDense & PixelGen & PixelDense & PixelGen & PixelDense & PixelGen & PixelDense \\
\midrule
\multirow{6}{*}{COCO val} & \multirow{3}{*}{Real panoptic} & 0.5 & 23.21 & \textbf{31.32} & 26.06 & \textbf{35.68} & 18.91 & \textbf{24.75} & 37.80 & \textbf{45.59} \\
 & & 0.7 & 15.70 & \textbf{21.31} & 17.07 & \textbf{23.63} & 13.65 & \textbf{17.80} & 30.67 & \textbf{36.70} \\
 & & 0.9 & 7.97 & \textbf{9.57} & 7.90 & \textbf{9.67} & 8.08 & \textbf{9.42} & 22.22 & \textbf{24.20} \\
\cmidrule(lr){2-11}
 & \multirow{3}{*}{Pseudo orig.} & 0.5 & 29.34 & \textbf{40.99} & 31.23 & \textbf{43.66} & 26.49 & \textbf{36.95} & 38.77 & \textbf{48.19} \\
 & & 0.7 & 18.86 & \textbf{26.37} & 19.66 & \textbf{27.57} & 17.65 & \textbf{24.55} & 30.69 & \textbf{37.26} \\
 & & 0.9 & 9.18 & \textbf{11.25} & 8.71 & \textbf{10.79} & 9.89 & \textbf{11.94} & 21.97 & \textbf{24.10} \\
\midrule
\multirow{3}{*}{Flickr30K} & \multirow{3}{*}{Pseudo orig.} & 0.5 & 19.96 & \textbf{30.98} & 17.53 & \textbf{28.88} & 23.40 & \textbf{33.90} & 19.94 & \textbf{29.42} \\
 & & 0.7 & 10.86 & \textbf{16.69} & 7.74 & \textbf{13.69} & 15.33 & \textbf{20.94} & 14.56 & \textbf{19.03} \\
 & & 0.9 & 4.05 & \textbf{5.69} & 1.87 & \textbf{3.55} & 7.09 & \textbf{8.76} & 9.64 & \textbf{11.10} \\
\bottomrule
\end{tabular}%
}
\end{table}

\begin{table}[t]
\centering
\caption{\textbf{Marigold v1.1 depth AbsRel against pseudo references, mean and median.} Mean and median per-image AbsRel after closed-form per-image scale and shift fitting, lower is better. The median AbsRel agrees with the mean on every cell, so the comparison is not driven by a few badly aligned outliers.}
\label{tab:appendix_marigold_depth}
\setlength{\tabcolsep}{4pt}
\begin{tabular}{l c | c c | c c}
\toprule
\multirow{2}{*}{Dataset} & \multirow{2}{*}{$\tau$} & \multicolumn{2}{c|}{Mean AbsRel $\downarrow$} & \multicolumn{2}{c}{Median AbsRel $\downarrow$} \\
 & & PixelGen & PixelDense & PixelGen & PixelDense \\
\midrule
\multirow{3}{*}{COCO val}  & 0.5 & 1.5805 & \textbf{1.0639} & 1.0559 & \textbf{0.6752} \\
                            & 0.7 & 2.0919 & \textbf{1.6286} & 1.5393 & \textbf{1.1179} \\
                            & 0.9 & 2.8269 & \textbf{2.6105} & 2.2362 & \textbf{2.0290} \\
\midrule
\multirow{3}{*}{Flickr30K} & 0.5 & 1.6061 & \textbf{1.0257} & 1.1443 & \textbf{0.6770} \\
                            & 0.7 & 2.2186 & \textbf{1.6639} & 1.7021 & \textbf{1.2144} \\
                            & 0.9 & 3.0102 & \textbf{2.7662} & 2.4926 & \textbf{2.3138} \\
\bottomrule
\end{tabular}
\end{table}

\begin{table}[t]
\centering
\caption{\textbf{Marigold v1.1 surface-normal angular error against pseudo references, mean and per-image-median statistics.} Mean pixelwise angular error in degrees, median per-image angular error, and mean of per-image medians. Lower is better. Per-image medians track the means on every cell, so the comparison is not driven by tail pixels.}
\label{tab:appendix_marigold_normals}
\setlength{\tabcolsep}{4pt}
\begin{tabular}{l c | c c | c c | c c}
\toprule
\multirow{2}{*}{Dataset} & \multirow{2}{*}{$\tau$} & \multicolumn{2}{c|}{Mean ($^\circ$) $\downarrow$} & \multicolumn{2}{c|}{Median per-image ($^\circ$) $\downarrow$} & \multicolumn{2}{c}{Mean of per-image medians ($^\circ$) $\downarrow$} \\
 & & PixelGen & PixelDense & PixelGen & PixelDense & PixelGen & PixelDense \\
\midrule
\multirow{3}{*}{COCO val}  & 0.5 & 21.80 & \textbf{17.84} & 20.79 & \textbf{16.48} & 15.25 & \textbf{11.76} \\
                            & 0.7 & 26.90 & \textbf{23.48} & 26.61 & \textbf{22.80} & 19.97 & \textbf{16.50} \\
                            & 0.9 & 33.20 & \textbf{31.94} & 33.70 & \textbf{32.36} & 26.50 & \textbf{24.93} \\
\midrule
\multirow{3}{*}{Flickr30K} & 0.5 & 22.45 & \textbf{18.25} & 21.03 & \textbf{16.88} & 16.27 & \textbf{12.26} \\
                            & 0.7 & 28.12 & \textbf{24.62} & 27.55 & \textbf{23.59} & 21.88 & \textbf{17.84} \\
                            & 0.9 & 34.97 & \textbf{33.88} & 35.39 & \textbf{34.52} & 29.18 & \textbf{27.98} \\
\bottomrule
\end{tabular}
\end{table}

\paragraph{What the diagnostic says about knowledge transmission.}
The diagnostic is consistent with the structural argument of \Cref{sec:method_pixeldense}. An unfactored objective sums four teacher cosine gradients on the same denoiser hidden state without reserving semantic and geometric subspaces. This can align per-step loss deltas across the bank and leave slower teachers behind on the longer horizon, which is why the naive run shows both higher cross-teacher step-delta cosine and a larger semantic-versus-geometric reduction gap. Factoring the objective across the semantic-versus-geometric supervision split lets the per-step gradient match the active teacher group with less drag from the other group, which is what the lower cross-stream step-delta cosine measures. The separated update shows up as larger and more uniform per-teacher loss reductions over $10{,}000$ steps, which we read as more even knowledge transmission from the four frozen teachers into the denoiser. The training-time advantage in \Cref{tab:repa_dynamics_summary,tab:repa_dynamics_progress} is the mechanism behind the GenEval gap reported in \Cref{tab:teacher_compose}.

\subsection{Per-Backbone Geometry Preservation Numbers}
\label{sec:appendix_perbackbone_geom}

\Cref{tab:geometry_reconstruction} averages the OneFormer panoptic scores across the Swin-L and DiNAT-L backbones. \Cref{tab:appendix_oneformer_swin,tab:appendix_oneformer_dinat} report the two backbones separately and break PQ into PQ$_\text{th}$ for thing classes and PQ$_\text{st}$ for stuff classes. The two backbones disagree by at most $0.005$ on PQ and mIoU per cell, so the averaged numbers in \Cref{tab:geometry_reconstruction} are not driven by either backbone in particular. \Cref{tab:appendix_marigold_depth,tab:appendix_marigold_normals} report the dense-geometry probes with both mean and median statistics; the median AbsRel and the mean of per-image median angular errors agree with the means on every dataset and $\tau$ cell, so the ranking in \Cref{tab:geometry_reconstruction} is not driven by tail outliers.

\begin{figure}[t]
    \centering
    \includegraphics[width=\textwidth]{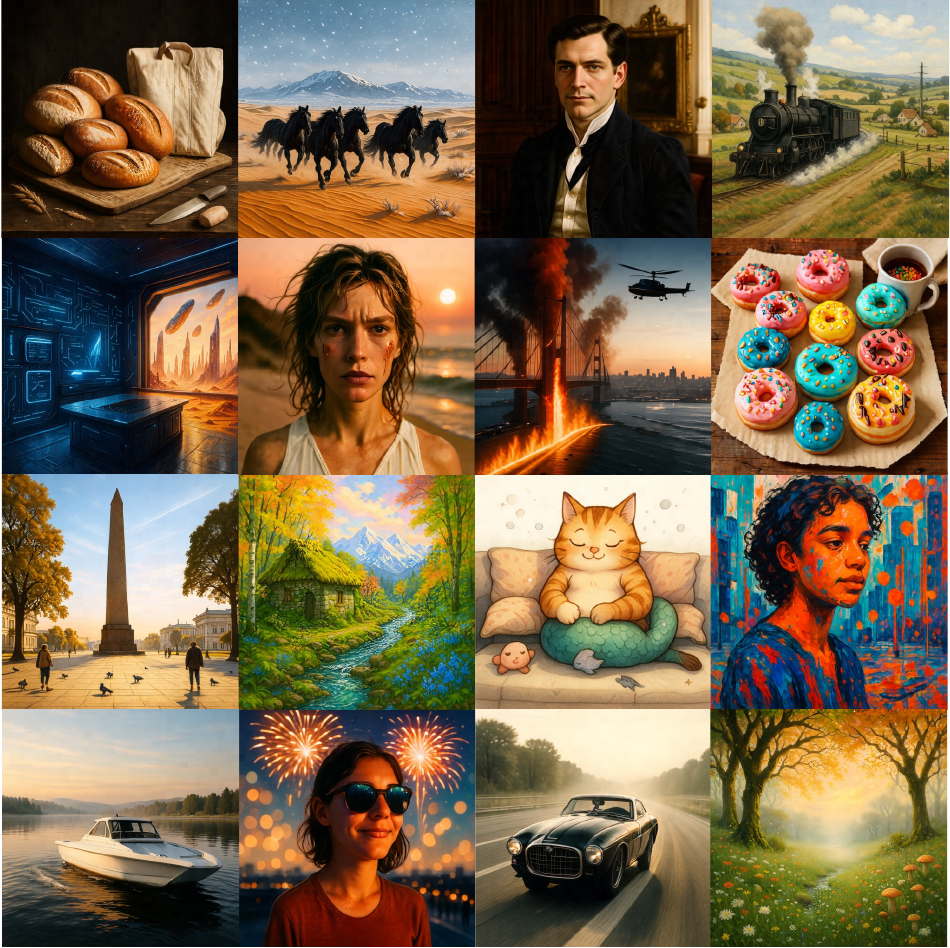}
    \caption{\textbf{Additional PixelDense samples on the PixelGen-XXL backbone at $512 \times 512$.} Object-centered photographs, painterly portraits, atmospheric landscapes, and stylized fantasy scenes. The dense teachers are dropped at inference, and sampling matches the protocol of \Cref{sec:exp_setup}.}
    \label{supp:supp_qualitative}
\end{figure}

\begin{figure}[t]
    \centering
    \includegraphics[width=\textwidth]{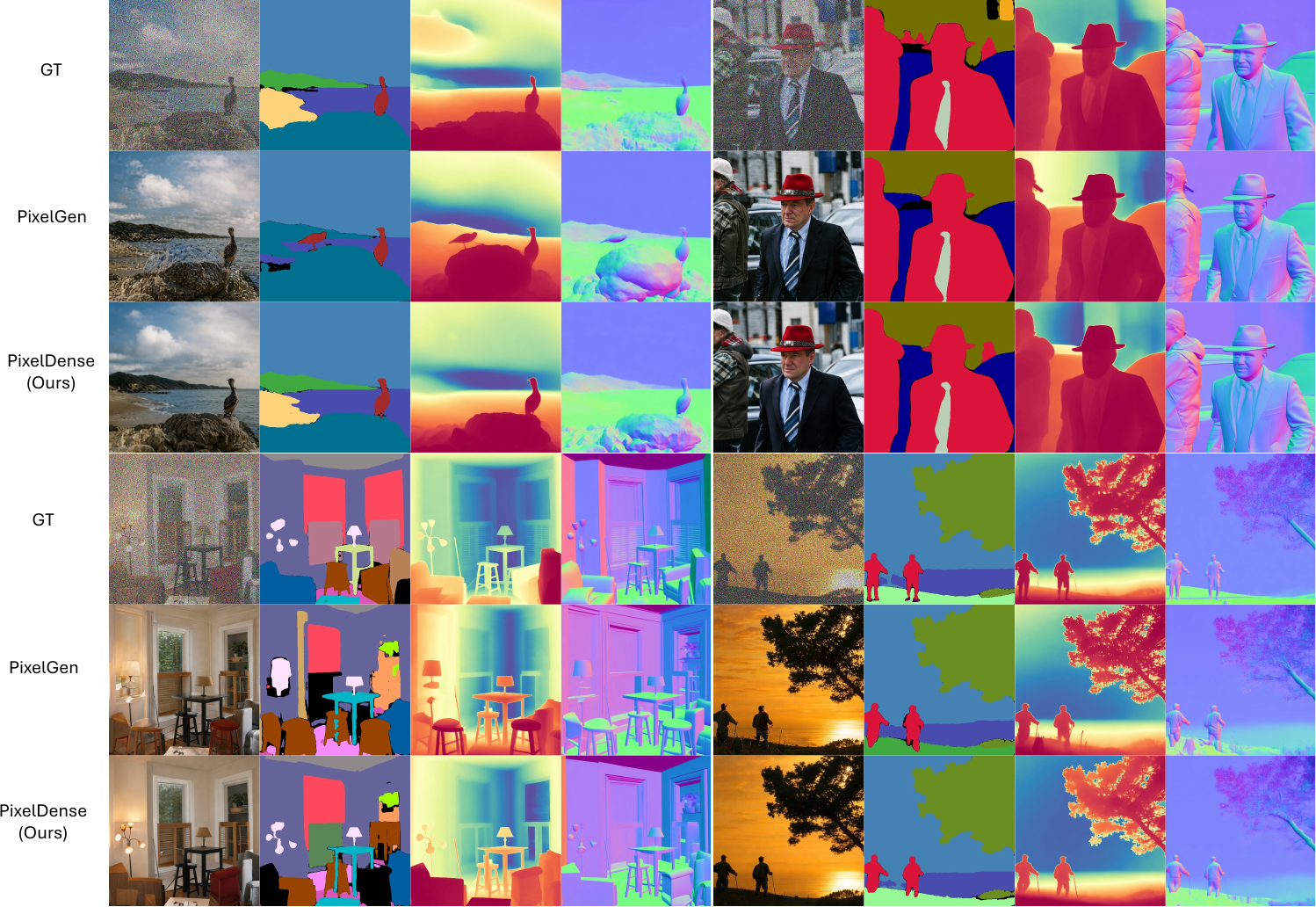}
    \caption{\textbf{Additional partial-noise reconstruction comparisons between PixelGen and PixelDense.} Each row shows the noised image at the indicated $\tau$, the two reconstructions, and the panoptic, depth, and surface-normal probes from OneFormer~\citep{jain2023oneformer} and Marigold~\citep{ke2024repurposing}. PixelDense reconstructions track object boundaries, depth ordering, and surface orientation more faithfully than PixelGen at every $\tau$.}
    \label{supp:supp_geometric}
\end{figure}

\subsection{HPS v2.1 Per-Style Breakdown}
\label{sec:appendix_hps}

\Cref{tab:hps_style} reports the HPS v2.1 scores for each of the four $800$-prompt style splits of \citet{wu2023human}. PixelDense matches or exceeds the matched PixelGen DINOv2-only fine-tune on every split and on the four-split average, with the largest gain on Paintings and Concept-art. The released PixelGen-XXL T2I checkpoint is included for reference.

\begin{table}[h]
\centering
\caption{\textbf{HPS v2.1 per-style breakdown across the four prompt categories of \citet{wu2023human}.} Each split contains $800$ prompts, and the rightmost column averages the four splits. Higher is better, and bold marks PixelDense entries that improve over the matched PixelGen fine-tune.}
\label{tab:hps_style}
\begin{tabular}{l | c c c c | c}
\toprule
Model & Anime & Concept-art & Paintings & Photo & Average \\
\midrule
PixelGen-XXL T2I, released checkpoint & 0.292 & 0.278 & 0.269 & 0.284 & 0.281 \\
DeCo T2I, released checkpoint         & 0.285 & 0.267 & 0.259 & 0.277 & 0.272 \\
\midrule
PixelGen + DINOv2-only fine-tune  & 0.290 & 0.277 & 0.268 & 0.285 & 0.280 \\
\textbf{PixelGen + PixelDense, ours}  & \textbf{0.292} & \textbf{0.280} & \textbf{0.272} & 0.284 & \textbf{0.282} \\
\bottomrule
\end{tabular}
\end{table}

\subsection{Additional Qualitative Visualization}
\label{sec:appendix_qualitative}

\Cref{supp:supp_qualitative} extends the prompt range of the main-paper teaser, covering object-centered photographs, painterly portraits, atmospheric landscapes, and stylized fantasy scenes at $512 \times 512$. All samples use the $25$-step Adam LM solver and classifier-free guidance scale $4.0$ of \Cref{sec:exp_setup}, and the dense teachers are absent at inference. \Cref{supp:supp_geometric} extends \Cref{fig:geometry_preservation} with additional partial-noise reconstruction comparisons between PixelGen and PixelDense at the same noise fractions as \Cref{tab:geometry_reconstruction}.

\section{Asset Licenses and Use}
\label{sec:appendix_licenses}

We use the public assets listed in \Cref{tab:asset_licenses} without modification, in accordance with their published licenses. We did not collect new data, did not redistribute any of these assets.

\begin{table}[h]
\centering
\caption{\textbf{Public assets used in this paper, with the license under which each is released.} All assets are used without modification. The training corpus and evaluation prompt sets are used through their official releases.}
\label{tab:asset_licenses}
\setlength{\tabcolsep}{6pt}
\begin{tabular}{l l l}
\toprule
Role & Asset & License \\
\midrule
Pixel backbone           & PixelGen-XXL~\citep{ma2026pixelgen}             & Apache 2.0 use \\
Pixel backbone           & DeCo-T2I~\citep{ma2025deco}                     & Apache 2.0 use \\
\midrule
Semantic teacher         & DINOv2~\citep{oquab2023dinov2}                  & Apache 2.0 \\
Semantic teacher         & SAM2~\citep{ravi2024sam}                        & Apache 2.0 \\
Geometric teacher        & Depth Anything v2~\citep{yang2024depthv2}       & Apache 2.0 \\
Geometric teacher        & Metric3D v2~\citep{hu2024metric3d}              & BSD 2-Clause \\
\midrule
Panoptic probe           & OneFormer~\citep{jain2023oneformer}             & MIT \\
Panoptic backbone        & Swin-L~\citep{liu2021swin}                      & MIT \\
Panoptic backbone        & DiNAT-L~\citep{hassani2022dilated}              & MIT \\
Depth and normal probe   & Marigold v1.1~\citep{ke2024repurposing}         & Apache 2.0 \\
\midrule
Training corpus          & BLIP3-o-60K~\citep{chen2025blip3}               & Apache 2.0 \\
Evaluation               & GenEval~\citep{ghosh2023geneval}                & MIT \\
Evaluation               & DPG-Bench~\citep{hu2024ella}                    & Apache-2.0 \\
Evaluation               & HPS v2.1~\citep{wu2023human}                    & Apache 2.0 \\
Evaluation               & COCO 2017~\citep{lin2014microsoft}              & CC BY 4.0, Flickr terms \\
Evaluation               & Flickr30K~\citep{plummer2015flickr30k}          & research use \\
\bottomrule
\end{tabular}
\end{table}

\section{Broader Impact and Limitations}
\label{sec:appendix_broader_impact}

PixelDense improves the training-time supervision of pixel-space text-to-image diffusion models without changing the deployed sampler. Like other text-to-image generators, the resulting models can support creative tools and accessible visual prototyping, and can also be misused for fabricated imagery, identity-targeted content, or copyright-sensitive style mimicry. The teachers we add are perception models trained on natural images, so PixelDense inherits whatever biases are present in those teachers' training distributions on top of the biases of the pixel backbone. We did not change the training dataset of the backbone, so PixelDense does not introduce a new data-collection footprint, and the dense teachers are dropped at inference, so deployment cost and downstream filtering pipelines are unchanged.

The current alignment is applied at a single mid-network block of the denoiser. Multi-scale and multi-block alignment, scaling the backbone and the training corpus, and exploring teacher banks beyond the four perception models studied here are natural extensions. Image editing often requires preserving instance identity, scene layout, object boundaries, depth, and surface orientation while changing only the requested content. Under SDEdit, which uses the trained denoiser as is, PixelDense keeps more of the source background and layout (\Cref{tab:pie_edit}). Other editing methods, such as inversion-based and instruction-based editing, add inversion or extra training on top of the denoiser. We leave them to future work, along with metrics for identity and 3D geometry after editing.

\end{document}